\documentclass[runningheads]{llncs}

\usepackage[mobile]{eccv}

\usepackage{eccvabbrv}

\usepackage{arydshln}
\usepackage{graphicx}
\usepackage{booktabs}
\usepackage{multirow}
\usepackage{kotex}
\usepackage{tabularx}
\usepackage{ulem}
\usepackage{makecell}
\usepackage{array}
\usepackage{placeins}
\usepackage{pgffor}

\usepackage[accsupp]{axessibility}

\usepackage[pagebackref,breaklinks,colorlinks]{hyperref}

\usepackage{orcidlink}

\begin{document}

\title{An Image Grid Can Be Worth a Video: Zero-shot Video Question Answering Using a VLM}

\titlerunning{An Image Grid Can Be Worth a Video}

\author{Wonkyun Kim\inst{1}, 
Changin Choi\inst{2} \and
Wonseok Lee\inst{2} \and
Wonjong Rhee\inst{1,2}, Fellow, IEEE}

\authorrunning{Wonkyun Kim et al.}

\institute{Department of Intelligence and Information, Seoul National University \and Interdisciplinary Program in Artificial Intelligence, Seoul National University\\
\email{\{wonkyunkim.sj, ci2015.choi,  dnjstjr1017, wrhee\}@snu.ac.kr}
} 

\maketitle

\begin{abstract}
Stimulated by the sophisticated reasoning capabilities of recent Large Language Models~(LLMs), a variety of strategies for bridging video modality have been devised. A prominent strategy involves Video Language Models~(VideoLMs), which train a learnable interface with video data to connect advanced vision encoders with LLMs. Recently, an alternative strategy has surfaced, employing readily available foundation models, such as VideoLMs and LLMs, across multiple stages for modality bridging. In this study, we introduce a simple yet novel strategy where only a single Vision Language Model~(VLM) is utilized. Our starting point is the plain insight that a video comprises a series of images, or frames, interwoven with temporal information. The essence of video comprehension lies in adeptly managing the temporal aspects along with the spatial details of each frame. Initially, we transform a video into a single composite image by arranging multiple frames in a grid layout. The resulting single image is termed as an \textit{image grid}. This format, while maintaining the appearance of a solitary image, effectively retains temporal information within the grid structure. Therefore, the image grid approach enables direct application of a single high-performance VLM without necessitating any video-data training. Our extensive experimental analysis across ten zero-shot video question answering benchmarks, including five open-ended and five multiple-choice benchmarks, reveals that the proposed \textit{Image Grid Vision Language Model}~(IG-VLM) surpasses the existing methods in nine out of ten benchmarks.\footnote{Our code is are available at: \url{https://github.com/imagegridworth/IG-VLM}}
    \keywords{Image Grid \and Video Representation \and Vision Language Model \and Video Question Answering}
\end{abstract}
\section{Introduction}
\label{sec:intro}

\begin{figure}[tb]
  \centering
  \includegraphics[width=\linewidth]{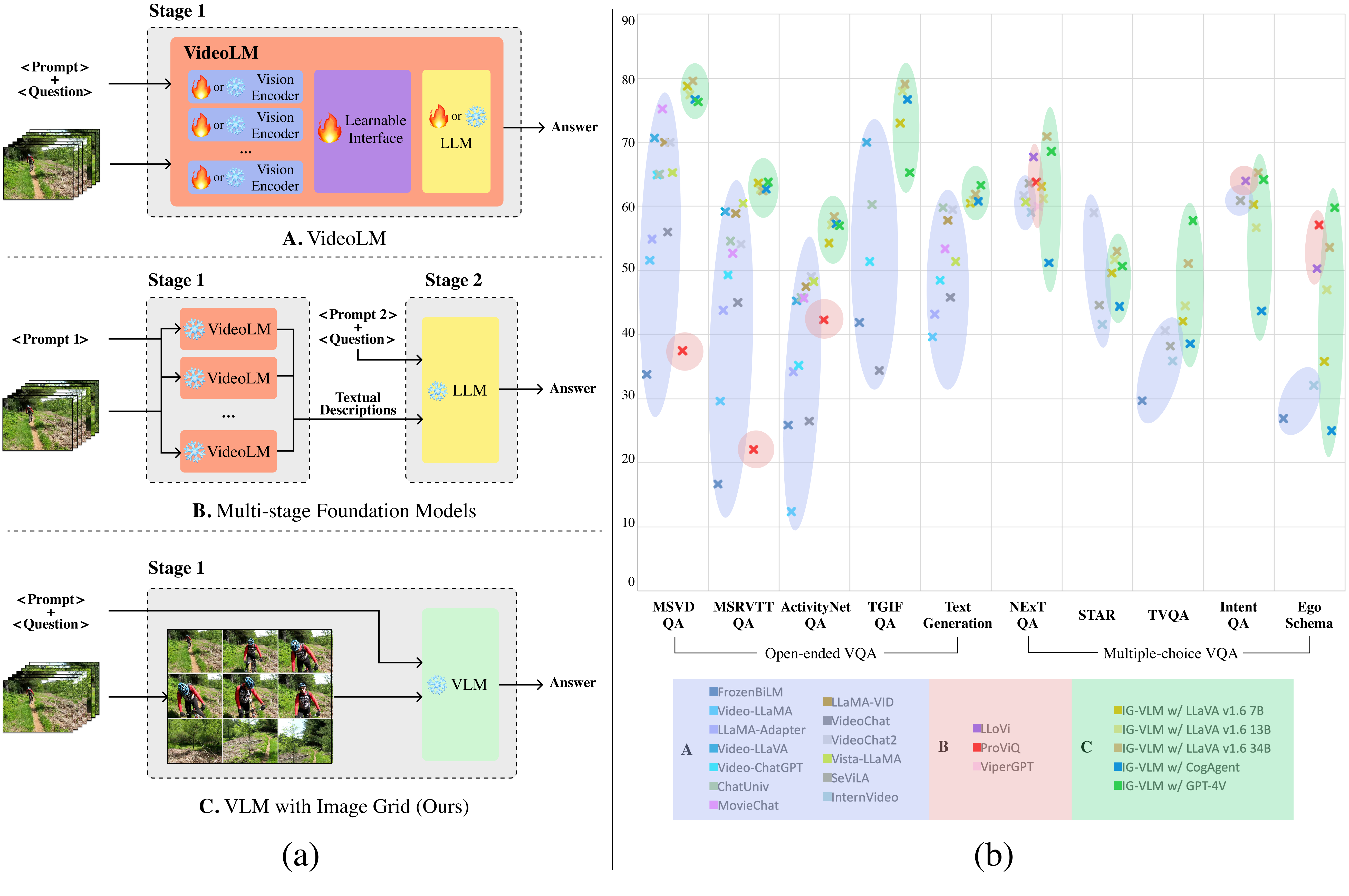}
  \caption{Zero-shot video question answering: (a)~Three approaches for integrating video content into LLMs. (b)~Performance summary for ten benchmarks. In contrast to many of the previous studies, we conducted an extensive experiment that encompasses the most widely recognized benchmarks for both open-ended and multiple-choice zero-shot Video Question Answering (VQA). For the text generation benchmark, performance was normalized to a maximum score of 100.
  }
  \label{fig:overview}
\end{figure}

Large Language Models~(LLMs) have demonstrated exceptional reasoning capabilities~\cite{Brown2020LanguageMA,Chowdhery2022PaLMSL,Anil2023PaLM2T,achiam2023gpt,Touvron2023LLaMAOA}. Because the modality of LLMs is language, the predominant mode for humans to exchange information and express requests, the success of LLMs has enabled a broad array of human tasks to be solved without task-specific models or training.
Vision Language Model~(VLM) is the most notable extension of LLM. By bridging vision modality into LLMs, VLMs have forged a highly effective linkage between visual data and the reasoning power of LLMs~\cite{Koyejo2022FLAMINGONIPS,wang2022simvlm,Yu2022CoCaCC,Li2022BLIPBL,Li2023BLIP2BL,Ye2023mPLUGOwlME,Dai2023InstructBLIPTG,liu2023llava}. This bridging allows the reasoning capabilities of LLMs to be effectively utilized across visual and textual data simultaneously.
Following the success of VLMs, researchers are actively investigating ways to integrate LLMs with other modalities such as audio~\cite{huang2023audiogpt,deshmukh2023pengi,Tang2023SALMONNTG,Qwen-Audio} and video~\cite{Li2023VideoChatCV,zhang2023VideoLLAMA,Maaz2023VideoChatGPTTD,Lin2023VideoLLaVALU}. 
In this study, we focus on the integration of video modality, with a specific emphasis on zero-shot Video Question Answering (VQA). 

Various strategies for integrating video content into LLMs have been explored, but no single method has yet emerged as the definitive solution. 
From the perspective of passing information to LLMs, existing strategies can be categorized into two groups, as illustrated in A and B of \cref{fig:overview}(a). Type A demonstrates \textit{VideoLM} approach, where both \textit{text tokens} and \textit{video tokens} serve as inputs to the LLM~\cite{Yang2022FrozenBiLM,Maaz2023VideoChatGPTTD,Lin2023VideoLLaVALU,Li2023LLaMAVIDAI,Luo2023ValleyVA,Li2023VideoChatCV,Ma2023VistaLLaMARV,li2023mvbench,zhang2023VideoLLAMA,Song2023MovieChatFD,Jin2023ChatUniViUV,Wang2023VLAPEV,yu2023self}.
In this popular approach, a learnable interface is trained with video data. Type B illustrates the approach of \textit{multi-stage foundation models}, which relies solely on \textit{text tokens} as LLM inputs~\cite{Choudhury2023ZeroShotVQ,Zhang2023ASL}. Here, video information undergoes pre-processing in stage 1, typically with off-the-shelf VideoLMs, to extract textual descriptions that are then fed into the reasoning LLM in the final stage.

The VideoLM approach leverages advanced vision encoders and LLMs, focusing primarily on their integration through a learnable interface. Although it adopts learning strategies akin to those used in VLMs, there remains room for enhancement to match the efficacy of VLMs. Currently, two notable challenges of VideoLMs are the limited availability of video data~\cite{yang2023vid2seq,yang2024vidchapters} and the constraint of encapsulating the full extent of video content into a restricted number of video tokens. 
The approach of multi-stage foundation models aligns with the recent trend of utilizing a suite of foundation models while avoiding task-specific training. This approach typically depends on pre-trained VideoLMs in the first stage, complemented by a variety of additional techniques aimed at outperforming the VideoLM-only strategy. While leveraging the reasoning capabilities of LLM, converting video into textual descriptions does not fully provide sufficient information about the video~\cite{Guo2022FromIT,Shao_2023_CVPR}. This process creates a modality connection gap~\cite{ModalityGap,Wang2023FillingTI}, essentially presenting the video to a \textit{blind} LLM that lacks innate understanding of video tasks, relying solely on the description. Furthermore, as it currently stands, it remains uncertain whether the strategy of multi-stage foundation models can definitively outperform VideoLMs. 
\Cref{fig:overview}(b) provides a summary of the experimental results across ten benchmarks. 
Here, it remains ambiguous whether type B strategies surpass type A, particularly in the context of open-ended VQA benchmarks.

To address the limitations identified in types A and B, we introduce a straightforward yet effective method centered on a single VLM, as shown in type C of \cref{fig:overview}(a). Generally, VLMs are designed to process only \textit{text tokens} and \textit{image tokens}. The latter only encapsulate spatial information, lacking the capacity to convey temporal aspects. Because of this difference between image tokens and video tokens, VLMs have been commonly considered infeasible for video understanding tasks. To overcome the problem, we propose to convert a video into a single composite image that arranges sampled frames from a video in a grid layout, thereby making it possible to encapsulate temporal information within a single image. This methodology offers several benefits: firstly, it allows for the exploitation of high-performance VLMs in video analysis; secondly, it obviates the need for video data because we completely avoid any video training by relying on pre-trained and frozen VLMs; and thirdly, it simplifies the process by eliminating the necessity for multi-stage foundation models. Remarkably, our straightforward approach outperforms the existing state-of-the-art methods in nine out of ten benchmarks as long as a proper VLM is chosen for each benchmark -- five out of five for open-ended VQA benchmarks and four out of five for multiple-choice VQA benchmarks. The proposed method is named as IG-VLM, which stands for \textit{Image Grid Vision Language Model}.
\section{Related Works}

\subsection{VideoLM}
A VideoLM integrates video data and language models by incorporating learnable interfaces that jointly learn the spatial and temporal aspects of video. Commonly, a learnable interface utilizes a projection network~\cite{Yang2022FrozenBiLM,Maaz2023VideoChatGPTTD,Lin2023VideoLLaVALU,Li2023LLaMAVIDAI,Luo2023ValleyVA}, an inter-modality attention~\cite{Li2023VideoChatCV,Ma2023VistaLLaMARV,li2023mvbench}, or a modality perceiver~\cite{zhang2023VideoLLAMA,Song2023MovieChatFD,Jin2023ChatUniViUV,Wang2023VLAPEV,yu2023self}. 
These interfaces serve to bridge the spatial and temporal information of videos with the processing capabilities of LLMs, by converting video content into video tokens that LLMs can efficiently interpret.

As a pioneering work, FrozenBiLM~\cite{Yang2022FrozenBiLM} combined a frozen vision encoder with BiLM for efficient video processing. When compared to the preceding methods, it achieved an impressive zero-shot performance in VQA.
Following this, Video-ChatGPT~\cite{Maaz2023VideoChatGPTTD} introduced video instruction tuning with the creation of high-quality instruction data. In the work, video-based text generation was proposed as a benchmark.
VideoChat~\cite{Li2023VideoChatCV} leveraged cross-attention to compress video tokens with user queries and dialogue context. 
VideoChat2~\cite{li2023mvbench} further refined this process through a multi-stage bootstrapping approach, focusing on modality alignment and instruction tuning. In addition, it gathered a wide array of high-quality video data and transformed it into video instruction tuning data.
Valley~\cite{Luo2023ValleyVA} employed temporal modeling and visual encoding to enhance video processing, through frame sampling and feature aggregation. 
Video-LLaVA~\cite{Lin2023VideoLLaVALU} used a pre-aligned encoder for both images and videos, enabling shared projections and joint training in image and video tasks.

Long videos are characterized by high computational complexity and memory demands. 
For the approaches that represent the entire long video with video tokens, the inherent challenges in joint spatial and temporal modeling make it particularly difficult to understand the complexities of spatial detail and temporal context thoroughly.
To efficiently manage long videos, VideoLMs employed a variety of advanced temporal modeling techniques.
MovieChat~\cite{Song2023MovieChatFD} employed a unique memory mechanism in transformers and merged similar frames, reducing computational complexity and memory usage.
Chat-UniVi~\cite{Jin2023ChatUniViUV} introduced a unified model for images and videos using dynamic token merging with k-NN to condense spatial and temporal tokens.
LLaMA-VID~\cite{Li2023LLaMAVIDAI} adopted a dual token to efficiently compress the video token by distinguishing context and content.
Vista-LLaMA~\cite{Ma2023VistaLLaMARV} introduced EDVT-Attention and a sequential vision projector, focusing on visual tokens and reducing temporal tokens by sequentially merging tokens with a Q-former.
To streamline the processing of lengthy videos, some approaches prioritized the selection of crucial keyframes, thereby reducing the overall number of frames. 
SeViLA~\cite{yu2023self} focused on identifying and extracting keyframes pertinent to the questions and interpreted the video by converting these keyframes into video tokens. 
VLAP~\cite{Wang2023VLAPEV} employed end-to-end training to refine frame selection, using a learnable interface to extract query-related keyframes and enhancing the integration of video modality with keyframes.

\subsection{Multi-stage Foundation Models}
Building upon the VideoLM approach, another strategy emerged where multi-stage video modality bridging was adopted using off-the-shelf VideoLMs and LLMs. 
This multi-stage strategy converts video content into textual descriptions, typically using pre-trained VideoLMs, such that they can be passed to the LLM in the last stage. Representing a video as text tokens leverage LLMs' proficiency in text analysis, allowing them to interpret temporal information through the provided set of textual descriptions. 
VideoChat-Text~\cite{Li2023VideoChatCV} transformed video streams into textual descriptions, capturing various aspects of the video. 
LLoVi~\cite{Zhang2023ASL} introduced a simple yet effective LLM-based framework to answer long-range video questions. In their work, video captioners transformed videos into text descriptions with details for LLMs to summarize, and enhanced long-range video understanding.
While the two methods above transformed videos into textual descriptions for LLM interpretation, 
LLMs were also used for generating programs that can assist video analysis. 
ViperGPT~\cite{suris2023vipergpt} pioneered the utilization of code-generating LLMs, such as GPT-3 Codex~\cite{chen2021evaluating}. It leveraged a visual module API in response to textual queries and generated programs that analyze images or videos to provide answers for those queries.
ProViQ~\cite{Choudhury2023ZeroShotVQ} employed an LLM to generate Python programs for multi-stage procedural reasoning in zero-shot video queries, and used the programs to derive answers for the provided questions.
\FloatBarrier
\begin{figure}[t]
  \centering
  \includegraphics[width=\linewidth]{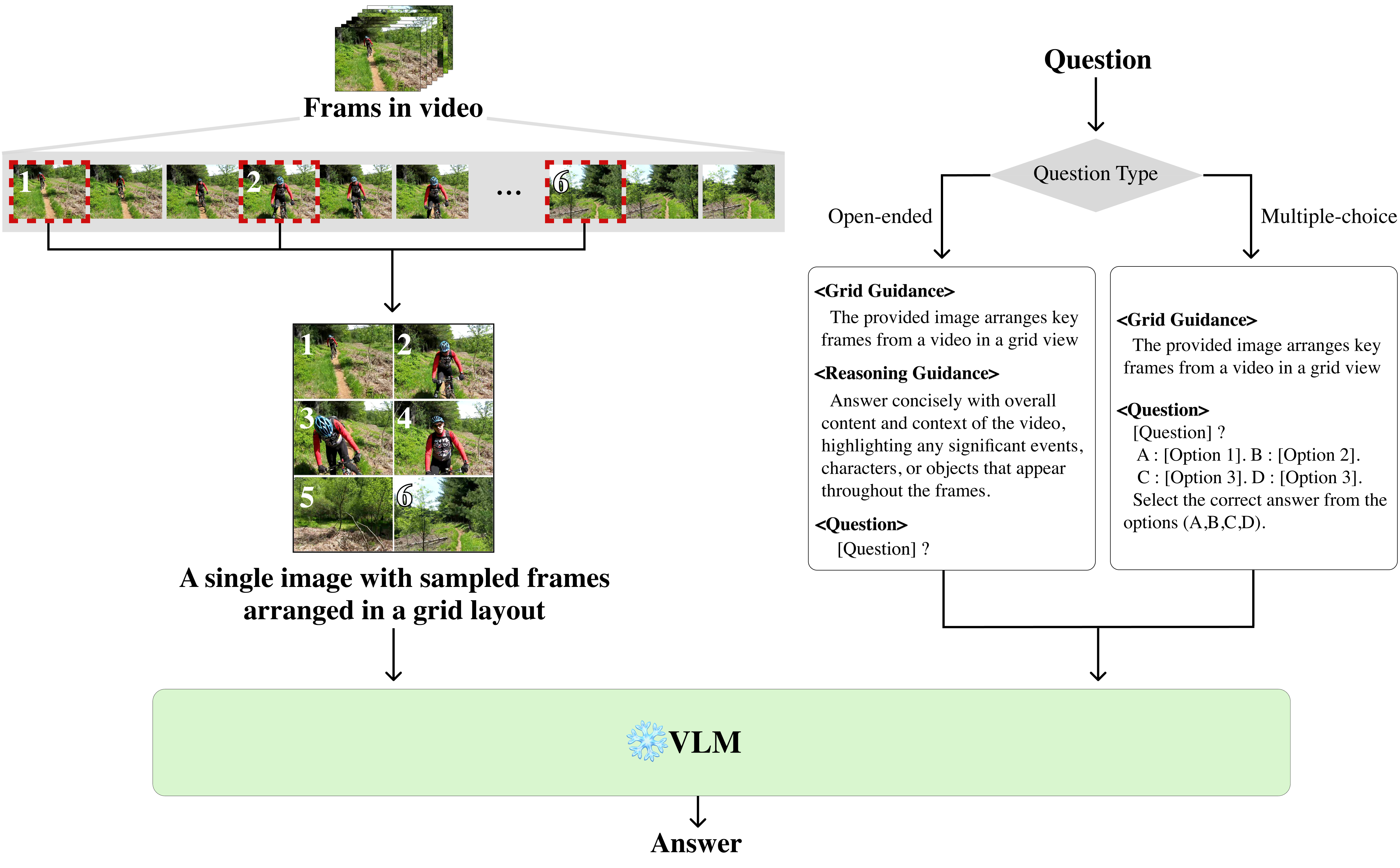}
  \caption{IG-VLM for zero-shot VQA: we transform a video into a single image comprising sampled frames arranged in a grid layout, which is then fed into a high-performance VLM alongside an appropriate prompt. The numbering included in the illustrations serves solely for explanatory purposes and is not incorporated into the actual methodology.
  }
  \label{fig:videogrid_overview}
\end{figure}

\section{Method}
\label{sec:method}

In this section, we introduce IG-VLM that leverages a single VLM for video question answering. Our method is visually illustrated in \cref{fig:videogrid_overview}. IG-VLM creates a video representation by organizing sampled frames into an image grid. Then, the image grid is fed into the VLM with the prompt, including grid guidance, reasoning guidance, and the question. Reasoning guidance is used only for open-ended questions. 

\subsection{Converting Video into Image Grid}
\label{subsec:grid}

To construct an image grid from a video, six frames are sampled from the provided video. The video sequence is uniformly partitioned into six intervals and the first frame in each interval is chosen to be included in the image grid. The six sampled frames from the six intervals are merged into a single image arranged in a grid layout. To make the resulting image grid closely approximate a square aspect ratio, the frames are arranged in a $3\times2$ layout. Specifically, frames are organized from left to right within each row, proceeding to the next row sequentially. We provide a comprehensive analysis of the image grid design in \cref{subsec:design_analysis_image_grid}. 

\subsection{Prompt for Image Grid}
\label{subsec:prompt}
Our prompt design incorporates three essential components: grid guidance prompt, reasoning guidance prompt, and the question prompt. The grid guidance prompt was designed to make VLM understand that an image grid contains a sequence of frames from a video in an ordered manner. 
The reasoning guidance prompt is adopted only for open-ended questions and it provides a guidance to VLM on how the answer should be derived considering image grid and grid guidance prompt. We provide the results of an ablation study in \cref{sec:prompt_design_ablation}.

\section{Experiments}
\label{sec:experiments}

\subsection{Datasets and Metrics}
\label{subsec:exp_setup}
We performed experiments over ten benchmarks. We explain the datasets and metrics first, and then provide the results.  

\subsubsection{Datasets:}
We evaluate IG-VLM across a broad spectrum of zero-shot video question-answer benchmarks. The benchmarks can be categorized into open-ended and multiple-choice depending on the question type. Open-ended VQA tasks require models to generate answers freely based on a video and its associated question. Multiple-choice VQA tasks present models with a set of predefined options and demand a correct selection from the options. For open-ended VQA, our evaluation includes benchmarks of MSVD-QA~\cite{xu2017video}, MSRVTT-QA~\cite{xu2017video}, ActivityNet-QA~\cite{yu2019activitynet}, TGIF-Frame-QA~\cite{Jang2017TGIFQATS}, and the Text Generation Performance benchmark~\cite{Maaz2023VideoChatGPTTD}. For multiple-choice VQA, we assess performance on NExT-QA~\cite{xiao2021next}, STAR~\cite{wu2021star}, TVQA~\cite{lei2018tvqa}, IntentQA~\cite{Li2023IntentQACV}, and EgoSchema~\cite{mangalam2023egoschema}. 

\subsubsection{Evaluation metric}
To evaluate open-ended VQA, we utilized GPT-assisted assessments proposed in Video-ChatGPT~\cite{Maaz2023VideoChatGPTTD}. The evaluation process enables us to thoroughly assess the accuracy and the correctness of the responses. For multiple-choice VQA, we focus on the accuracy based on model's ability to select the correct answer from the provided options. In the result tables, underlined numbers indicate the best performance reported prior to our work and bold numbers indicate the overall best performance. The overall best performance can occur for either prior work or our work.

\subsubsection{Implementation details}
We examined a total of five VLMs in our experiments to cover a range of pre-trained VLMs. They are CogAgent\cite{hong2023cogagent}, LLaVA v1.6\cite{liu2024llavanext} in three different sizes (7B, 13B, and 34B), and GPT-4V\cite{achiam2023gpt}. The aspect of LLM size is discussed in \cref{subsec:diss_LLM_size}.

\subsection{Open-ended VQA}
\label{subsec:open_vqa}

The experiment results for four open-ended VQA benchmarks are shown in \cref{tab:open_end_qa}. 
The performance of IG-VLM is superior to the previously reported best result, across all four open-ended benchmarks and regardless of the choice of VLM for IG-VLM. 
The only exception is IG-VLM based on GPT-4V for TGIF-QA dataset.
Particularly noteworthy is the performance on ActivityNet-QA, which has the longest average video length among the four benchmarks. The existing methods struggle with the long video benchmark.
IG-VLM demonstrates a significant improvement, achieving an enhancement of up to 9.3\% with LLaVA v1.6 34B.
Even if we consider only the two VLMs with 7B parameters, CogAgent achieves an improvement of 8.2\%.
It is encouraging to observe that IG-VLM can handle long videos better than the existing methods, despite its limitation of packing only six frames into the image grid.

The experiment results for the open-ended VQA benchmark of Text Generation are shown in \cref{tab:generative_benchmark}. 
For the overall performance that is summarized as the average score, IG-VLM is superior to the previously reported best performance regardless of the choice of VLM. This result is consistent with what we have observed in \cref{tab:open_end_qa}. 
When the five individual categories are investigated, however, IG-VLM does not necessarily outperform the previously reported best performance. 
In particular, the five IG-VLM models consistently perform worse than Chat-UniVi, VideoChat2, and LLaMA-VID, the three most recent models, under the category of DO~(Detail Orientation). This can be due to the limited number of frames that are packed into an image grid. 
In contrast, the five IG-VLM models consistently outperform the previous best performance for the categories of CI (Correctness of Information) and CO (Consistency). 
This consistent improvement can be due to the imperfection of modality bridging in the current VideoLMs. 
As for CU (Contextual Understanding) and TU (Temporal Understanding), that are key metrics for video understanding, IG-VLM generally exhibits a high performance together with the aforementioned three most recent models. The performance of other existing VideoLM methods is notably inferior.

\begin{table}[t]
\centering
\caption{Performance results for \textit{open-ended question} benchmarks. The performance of LLaMA-Adapter and VideoChat are from \cite{Li2023LLaMAVIDAI}.}
\label{tab:open_end_qa}
\scalebox{0.595}{
\begin{tabular}{cc|ccccc|cc|cc|cc|cc}
\toprule
\multicolumn{2}{c|}{\multirow{2}{*}{\textbf{Method}}} & \multirow{2}{*}{\makecell{\textbf{Vision}\\\textbf{Encoder}}} & \multirow{2}{*}{\makecell{\textbf{LLM}\\\textbf{Size}}} & \multicolumn{2}{c}{\textbf{Inference}} & \multirow{2}{*}{\makecell{\textbf{Video}\\\textbf{Trained}}} & \multicolumn{2}{c|}{\textbf{MSVD-QA}} & \multicolumn{2}{c|}{\textbf{MSRVTT-QA}} & \multicolumn{2}{c|}{\textbf{ActivityNet-QA}} & \multicolumn{2}{c}{\textbf{TGIF-QA}} \\ 

& & & & \textbf{Vision} & \textbf{LLM} & & \textbf{~Acc.~} & \textbf{Score} & \textbf{~Acc.~} & \textbf{Score} & \textbf{~~Acc.~~} & \textbf{Score} & \textbf{~Acc.~} & \textbf{Score} \\
\midrule
\multicolumn{2}{c|}{FrozenBiLM\cite{Yang2022FrozenBiLM}} & ViT-L & 1.3B & multiple & single & O & 33.8 & - & 16.7 & - & 25.9 & - & 41.9 & - \\
\multicolumn{2}{c|}{Video-LLaMA\cite{zhang2023VideoLLAMA}} & CLIP-G & 7B     & multiple & single& O & 51.6 & 2.5 & 29.6 & 1.8 & 12.4 & 1.1 & - & - \\
\multicolumn{2}{c|}{LLaMA-Adapter\cite{Zhang2023LLaMAAdapterEF}} & ViT-B & 7B    & multiple & single & O & 54.9 & 3.1 & 43.8 & 2.7 & 34.2 & 2.7 & - & - \\
\multicolumn{2}{c|}{Video-ChatGPT\cite{Maaz2023VideoChatGPTTD}} & ViT-L & 7B   & multiple & single & O & 64.9 & 3.3 & 49.3 & 2.8 & 35.2 & 2.7 & 51.4 & 3.0 \\
\multicolumn{2}{c|}{Video-LLaVA\cite{Lin2023VideoLLaVALU}} & ViT-L       & 7B   & multiple & single & O & 70.7 & 3.9 & 59.2 & 3.5 & 45.3 & 3.3 & \underline{70.0} & 4.0 \\
\multicolumn{2}{c|}{Chat-UniVi\cite{Jin2023ChatUniViUV}}   & ViT-L & 7B   & multiple & single & O & 65.0 & 3.6 & 54.6 & 3.1 & 45.8 & 3.2 & 60.3 & 3.4 \\
\multicolumn{2}{c|}{MovieChat\cite{Song2023MovieChatFD}}   & CLIP-G  & 7B   & multiple & single & O & \underline{75.2} & 3.8 & 52.7 & 2.6 & 45.7 & 3.4 & - & - \\
\multicolumn{2}{c|}{VideoChat\cite{Li2023VideoChatCV}}   & CLIP-G  & 7B   & multiple & single & O & 56.3 & 2.8 & 45.0 & 2.5 & 26.5 & 2.2 & 34.4 & 2.3 \\
\multicolumn{2}{c|}{VideoChat2\cite{li2023mvbench}}  & UMT-L          & 7B   & multiple & single & O & 70.0   & 3.9 & 54.1 & 3.3 & \underline{49.1} & 3.3 & - & - \\
\multicolumn{2}{c|}{Vista-LLaMA\cite{Ma2023VistaLLaMARV}} & CLIP-G  & 7B   & multiple & single & O & 65.3 & 3.6 &  \underline{60.5} & 3.3 & 48.3 & 3.3 & - & - \\ 
\multicolumn{2}{c|}{LLaMA-VID\cite{Li2023LLaMAVIDAI}}   & CLIP-G  & 13B   & multiple & single & O & 70.0   & 3.7 & 58.9 & 3.3 & 47.5 & 3.3 & - & - \\
\midrule
\multicolumn{2}{c|}{ProViQ\cite{Choudhury2023ZeroShotVQ}} & TimeSformer-L         & GPT-3.5   & multiple & multiple & X & 37.5 & - & 22.1 & - & 42.3 & - & - & - \\ 
\midrule
\multicolumn{1}{c|}{\multirow{5}{*}{IG-VLM}} &CogAgent\cite{hong2023cogagent} & CLIP-E & 7B   & single & single & X & 76.7 & 4.1 & 62.7 & 3.6 & 57.3 & 3.6 & 76.7 & 4.0\\
\multicolumn{1}{c|}{} & LLaVA v1.6\cite{liu2024llavanext}  & ViT-L  & 7B & single & single & X & 78.8 & 4.1 & 63.7 & 3.5 & 54.3 & 3.4 & 73.0 & 4.0\\
\multicolumn{1}{c|}{}                      &LLaVA v1.6\cite{liu2024llavanext} & ViT-L & 13B & single & single & X & 77.4 & 4.1 & 62.6 & 3.4 & 57.1 & 3.5 & 78.0 & 4.0\\
\multicolumn{1}{c|}{}                      &LLaVA v1.6\cite{liu2024llavanext} & ViT-L & 34B & single & single & X & \textbf{79.6} & 4.1 & 62.4 & 3.5 & \textbf{58.4} & 3.5 & \textbf{79.1} & 4.2 \\
\multicolumn{1}{c|}{}                      &GPT-4V\cite{achiam2023gpt} & Unknown & GPT-4 & single & single & X & 76.3 & 4.0 & \textbf{63.8} & 3.5 & 57.0 & 3.5 & 65.3 & 3.7 \\
\bottomrule
\end{tabular}
}
\end{table}

\begin{table}[t]
\centering
\caption{
Performance results for \textit{Text Generation} benchmark. The results include evaluation metrics such as CI (Correctness of Information), DO (Detail Orientation), CU (Contextual Understanding), TU (Temporal Understanding), and CO (Consistency).}
\label{tab:generative_benchmark}
\scalebox{0.74}{
\begin{tabular}{cc|ccccc|cccccc}
\toprule
\multicolumn{2}{c|}{\multirow{2}{*}{\textbf{Method}}}                    & \multirow{2}{*}{\makecell{\textbf{Vision}\\\textbf{Encoder}}} & \multirow{2}{*}{\makecell{\textbf{LLM}\\\textbf{Size}}} & \multicolumn{2}{c}{\textbf{Inference}} & \multirow{2}{*}{\makecell{\textbf{Video}\\\textbf{Trained}}} & \multirow{2}{*}{\textbf{CI}} & \multirow{2}{*}{\textbf{DO}} & \multirow{2}{*}{\textbf{CU}} & \multirow{2}{*}{\textbf{TU}} & \multirow{2}{*}{\textbf{CO}} & \multirow{2}{*}{\makecell{\textbf{Avg.}\\\textbf{Score}}} \\
           &                                &                                 &                      & \textbf{Vision}         & \textbf{LLM}          &                                 &                      &                     &                     &                     &                     &                                   \\ 
\midrule
\multicolumn{2}{c|}{Video-LLaMA~\cite{zhang2023VideoLLAMA}} & CLIP-G & 7B & multiple & single & O & 1.96              & 2.18  & 2.16 & 1.82  & 1.79  & 1.98 \\
\multicolumn{2}{c|}{LLaMA-Adapter~\cite{Zhang2023LLaMAAdapterEF}}                             & ViT-B                           & 7B                   & multiple       & single       & O                               & 2.03                 & 2.32                & 2.30                 & 1.98                & 2.15                & 2.16  \\
\multicolumn{2}{c|}{Video-ChatGPT~\cite{Maaz2023VideoChatGPTTD}}                               & ViT-L                           & 7B                   & multiple       & single       & O                               & 2.50                  & 2.57                & 2.69                & 2.16                & 2.20                 & 2.42                 \\
\multicolumn{2}{c|}{Chat-UniVi~\cite{Jin2023ChatUniViUV}}                               & ViT-L                           & 7B                   & multiple       & single       & O & 2.89 & 2.91 & 3.46 & \underline{\textbf{2.89}} & \underline{2.81} & \underline{2.99} \\
\multicolumn{2}{c|}{MovieChat~\cite{Song2023MovieChatFD}}                                  & CLIP-G                          & 7B                   & multiple       & single       & O                               & 2.76                 & 2.93                 & 3.01                & 2.24                & 2.42                & 2.67                 \\
\multicolumn{2}{c|}{VideoChat~\cite{Li2023VideoChatCV}}                                  & CLIP-G                          & 7B                   & multiple       & single       & O                               & 2.23                 & 2.50                 & 2.53                & 1.94                & 2.24                & 2.29                 \\
\multicolumn{2}{c|}{VideoChat2~\cite{li2023mvbench}}                                 & UMT-L                           & 7B                   & multiple       & single       & O                               & \underline{3.02}                 & 2.88                & 3.51                & 2.66                & \underline{2.81}                & 2.98                 \\
\multicolumn{2}{c|}{Vista-LLaMA~\cite{Ma2023VistaLLaMARV}}                                  & CLIP-G                          & 7B                   & multiple       & single       & O                               & 2.44                 & 2.64          & 3.18                & 2.26                & 2.31                & 2.57                 \\
\multicolumn{2}{c|}{LLaMA-VID~\cite{Li2023LLaMAVIDAI}}                                  & CLIP-G                          & 7B                   & multiple       & single       & O                               & 2.96                 & \underline{\textbf{3.00}}                   & \underline{3.53}                & 2.46                & 2.51                & 2.89                 \\
\midrule
\multicolumn{1}{c|}{\multirow{6}{*}{IG-VLM}} & CogAgent~\cite{hong2023cogagent}         & CLIP-E                          & 7B                   & single         & single       & X                         & 3.26 & 2.76 & 3.57 & 2.34 & 3.28 & 3.04 \\
\multicolumn{1}{c|}{} & LLaVA v1.6~\cite{liu2024llavanext}       & ViT-L                        & 7B                   & single         & single       & X  & 3.11 & 2.78 & 3.51 & 2.44 & 3.29 & 3.03 \\
\multicolumn{1}{c|}{}                        & LLaVA v1.6~\cite{liu2024llavanext}       & ViT-L                        & 13B                  & single         & single       & X  & 3.17 & 2.79 & 3.52 & 2.51 & 3.25 & 3.05 \\
\multicolumn{1}{c|}{}                        & LLaVA v1.6~\cite{liu2024llavanext}       & ViT-L                        & 34B                  & single         & single       & X  & 3.21 & 2.87 & 3.54 & 2.51 & \textbf{3.34} & 3.09 \\
\multicolumn{1}{c|}{}                        & GPT-4V\cite{achiam2023gpt} & Unknown                         & GPT-4                & single         & single       & X      & \textbf{3.40} & 2.80 & \textbf{3.61} & \textbf{2.89} & 3.13 & \textbf{3.17} \\ 
\bottomrule
\end{tabular}
}
\end{table}

\subsection{Multiple-choice VQA}
\label{subsec:multiple_vqa}

The experimental results for multiple-choice VQA are presented in \cref{tab:multi_benchmark}.
Although the results are less dominant compared to the exceptional performance in open-ended VQA, IG-VLM still secures the best performance in four out of five multiple-choice benchmarks, provided that an appropriate VLM is selected for each specific benchmark.
It is worth noting that multi-stage foundation model strategies tend to surpass VideoLM strategies in multiple-choice benchmarks. In particular,  when we exclude IG-VLM results, LLoVi achieves the best performance for NExT-QA and InternQA and ProViQ achieves the best performance for EgoSchema.
Performance of multi-stage foundation models for open-ended VQA, however, is generally not available except for the partial ProViQ results included in \cref{tab:open_end_qa}. The partial results are much worse than what VideoLMs can achieve. 
Therefore, we can conclude that VideoLMs tend to outperform multi-stage foundation models for open-ended VQAs and multi-stage foundation models tend to outperform VideoLMs for multiple-choice VQAs, if we consider the experiment results in \cref{tab:open_end_qa}, \cref{tab:generative_benchmark}, and \cref{tab:multi_benchmark} only. 
In contrast, IG-VLM excels in both categories of VQA benchmarks.

\begin{table}[t]
\centering
\caption{Performance results for \textit{multiple-choice question} benchmarks. Evaluation of TVQA was performed without subscripts. The EgoSchema performance of FrozenBiLM is reported in \cite{Zhang2023ASL}. The EgoSchema performance of InternVideo is sourced from \cite{Zhang2023ASL}, while the rest of the performance is from \cite{li2023mvbench}, respectively. * denotes a video captioner trained on Ego4D~\cite{Grauman2021Ego4DAT}, while EgoSchema is a multiple-choice VQA benchmark based on these Ego4D videos. † denote that report accuracy of EgoSchema test split.} 
\label{tab:multi_benchmark}
\scalebox{0.53}{
\begin{tabular}{cc|ccccc|cccc|ccccc|c|c|c}
\toprule
\multicolumn{2}{c|}{\multirow{3}{*}{\textbf{Method}}} & \multirow{3}{*}{\makecell{\textbf{Vision}\\\textbf{Encoder}}} & \multirow{3}{*}{\makecell{\textbf{LLM}\\\textbf{Size}}} & \multicolumn{2}{c}{\multirow{2}{*}{\textbf{Inference}}}                  & \multirow{3}{*}{\makecell{\textbf{Video}\\\textbf{Trained}}} & \multicolumn{4}{c|}{\textbf{NExT-QA}}                                                  & \multicolumn{5}{c|}{\textbf{STAR}}                                                                             & \multirow{3}{*}{\textbf{TVQA}} & \multirow{3}{*}{\textbf{IntentQA}} & \multirow{3}{*}{\makecell{\textbf{Ego-}\\\textbf{Schema}}} \\
\multicolumn{2}{c|}{}                        &                                 &                      & \multirow{2}{*}{\textbf{Vision}}  & \multirow{2}{*}{\textbf{LLM}} &                                 & \multirow{2}{*}{\textbf{Cas.}} & \multirow{2}{*}{\textbf{Tem.}} & \multirow{2}{*}{\textbf{Des.}} & \textbf{Avg.}  & \multirow{2}{*}{\textbf{Int.}} & \multirow{2}{*}{\textbf{Seq.}} & \multirow{2}{*}{\textbf{Pre.}} & \multirow{2}{*}{\textbf{Fea.}} & \textbf{Avg.}  &                       &                           &                            \\
\multicolumn{2}{c|}{}                        &                                 &                      &                         &                      &                                 &                       &                       &                       & \textbf{Acc.}  &                       &                       &                       &                       & \textbf{Acc.}  &                       &                           &                            \\ \midrule
\multicolumn{2}{c|}{FrozenBiLM~\cite{Yang2022FrozenBiLM}}                                 & VIT-L                           & 1B                   & multiple      & single        & O                               & -     & -     & -     & -         & -     & -     & -     & -     & -         & 29.7                  & -  & 26.9                       \\
\multicolumn{2}{c|}{InternVideo~\cite{Wang2022InternVideoGV}}                                & VIT-L                           & 1.3B                 & multiple      & single        & O                               & 48.0    & 43.4  & 65.1  & 59.1      & 43.8  & 43.2  & 42.3  & 37.4  & 41.6      & 35.9                  & - & 32.1                         \\
\multicolumn{2}{c|}{VideoChat2~\cite{li2023mvbench}}                                 & UMT-L                           & 7B                   & multiple      & single        & O                               & 61.9  & 57.4  & 69.9  & 61.7      & 58.4  & 60.9  & 55.3  & 53.1    & \underline{\textbf{59.0}}        & \underline{40.6}                  & - & -                         \\
\multicolumn{2}{c|}{Vista-LLaMA~\cite{Ma2023VistaLLaMARV}}                                 & CLIP-G                           & 7B                   & multiple      & single        & O                               & -  & -  & -  & 60.7      & -  & -  & -  & -    & -        & -                  & - & -                         \\
\multicolumn{2}{c|}{Sevilla~\cite{yu2023self}}                                    & ViT-L                & 2.85B                & multiple      & multiple      & O                               & 61.3  & 61.5  & 75.6  & 63.6      & 48.3  & 45.0    & 44.4  & 40.8  & 44.6      & 38.2                  & 60.9 & -                      \\
\midrule
\multicolumn{2}{c|}{ViperGPT~\cite{suris2023vipergpt}}                                   & ViT-G               & GPT-3          & multiple      & multiple      & X                               & -     & -     & -     & 60.0        & -     & -     & -     & -     & -         & -                     & - & -                         \\
\multicolumn{2}{c|}{ProViQ~\cite{Choudhury2023ZeroShotVQ}}                                     & TimeSformer-L                          & GPT-3.5              & multiple      & multiple      & X                               & -     & -     & -     & 64.6      & -     & -     & -     & -     & -         & -                     & - & \underline{57.1}*†                         \\
\multicolumn{2}{c|}{LLoVi~\cite{Zhang2023ASL}}                                      & ViT-L                    & GPT-3.5              & multiple      & multiple      & X                               & 69.5  & 61.0    & 75.6  & \underline{67.7}      & -     & -     & -     & -     & -         & -                     & \underline{64.0} & 50.3*†                 \\ 
\midrule
\multicolumn{1}{c|}{\multirow{5}{*}{IG-VLM}} & CogAgent~\cite{hong2023cogagent}         & CLIP-E                          & 7B                   & single        & single        & X                               & 52.3 & 47.3 & 65.9 & 52.8     & 39.8 & 47.4 & 40.5 & 43.6 & 44.4     & 38.6                  & 43.7        & 25.0            \\
\multicolumn{1}{c|}{}                        & LLaVA v1.6~\cite{liu2024llavanext}       & ViT-L                        & 7B                  & single        & single        & X                               & 63.1  & 57.3  & 74.9  & 63.1      & 49.3  & 50.1    & 48.4    & 48.8  & 49.6      & 42.1                  & 60.3  & 35.8                   \\
\multicolumn{1}{c|}{}                        & LLaVA v1.6~\cite{liu2024llavanext}       & ViT-L                        & 13B                  & single        & single        & X                               & 61.6  & 55.7  & 70.8  & 61.2      & 51.5  & 52.0    & 51.0    & 51.8  & 51.7      & 44.5                  & 59.7  & 47.0                   \\
\multicolumn{1}{c|}{}                        & LLaVA v1.6~\cite{liu2024llavanext}       & ViT-L                        & 34B                  & single        & single        & X                               & 72.2 & 65.7 & 77.3 & \textbf{70.9}      & 53.4 & 53.9 & 49.5 & 48.4 & 53.0     & 51.1                 & \textbf{65.3}      & 53.6               \\ 
\multicolumn{1}{c|}{}                        & GPT-4V~\cite{achiam2023gpt} & Unknown                         & GPT-4                & single        & single        & X                               & 69.8  & 63.6  & 74.7  & 68.6      & 50.6    & 52.3  & 44.9  & 46.9  & 50.7      & \textbf{57.8}                  &  64.2       & \textbf{59.8}           \\ 
\bottomrule
\end{tabular}
}
\end{table}

\section{Analysis and Ablations Studies}

In this section, we discuss the analysis that influenced the configuration of the IG-VLM image grid. Following that, we detail the outcomes of two ablation studies.

\subsection{Design analysis of image grid}
\label{subsec:design_analysis_image_grid}

We analyze three design elements of our image grid -- shape of the grid, ordering of the sampled frames, and the number of frames to include in an image grid. For the first two, we explore $N=4,6$ and $9$ in the analysis, where $N$ is the number of sampled frames from a video. For the number of frames, we investigate $N=4,6,9,12,16$ and $20$.

\subsubsection{Shape of the image grid:}
\label{subsec:shape_of_image_grid}
\begin{figure}[]
    \centering
    \begin{subfigure}{0.40\linewidth}
        \includegraphics[width=\linewidth]{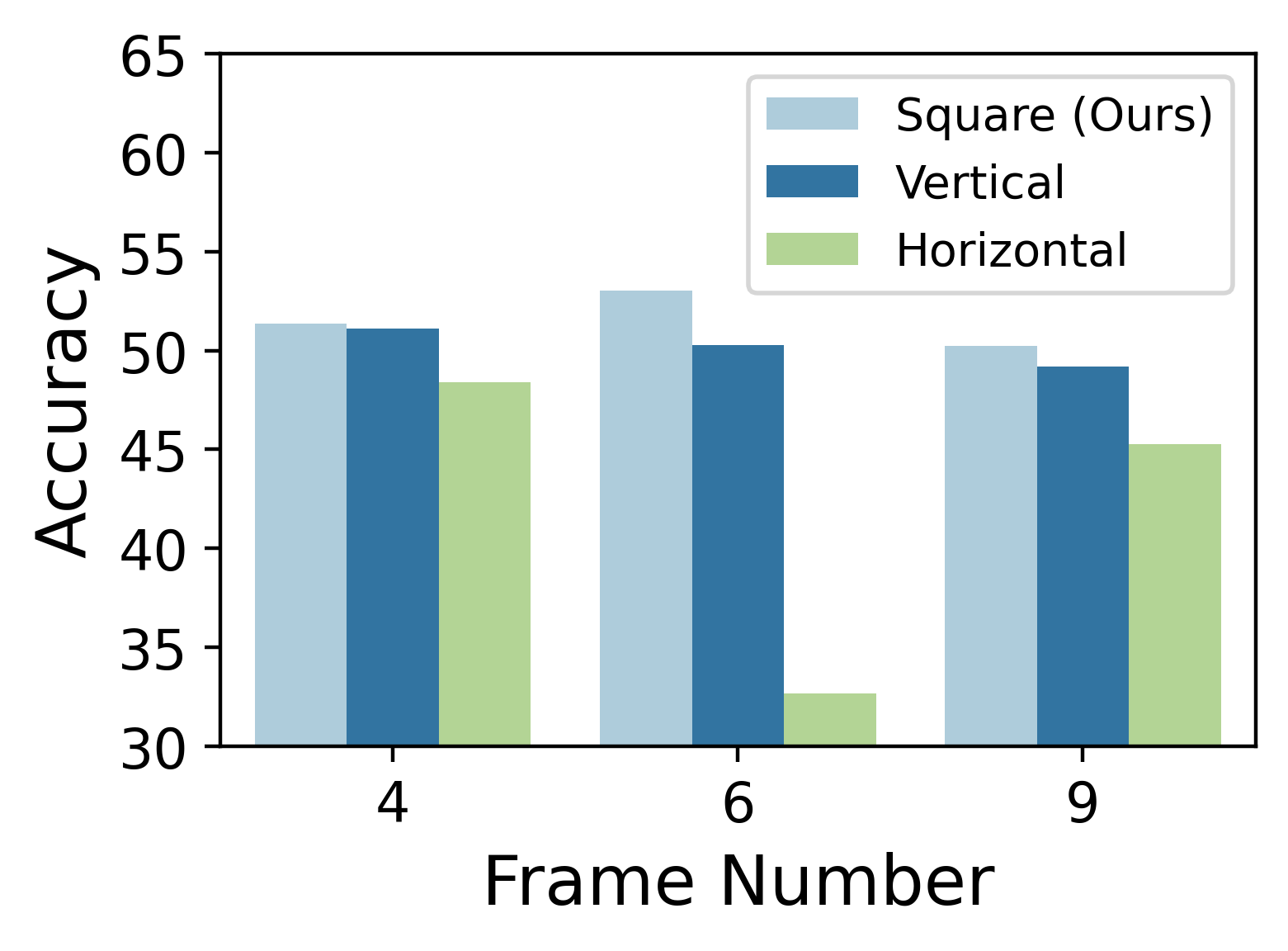}
        \caption{ActivityNet-QA}
        \label{fig:analysis_shape_and_ordering:shape1}
    \end{subfigure}
    \begin{subfigure}{0.40\linewidth}
        \includegraphics[width=\linewidth]{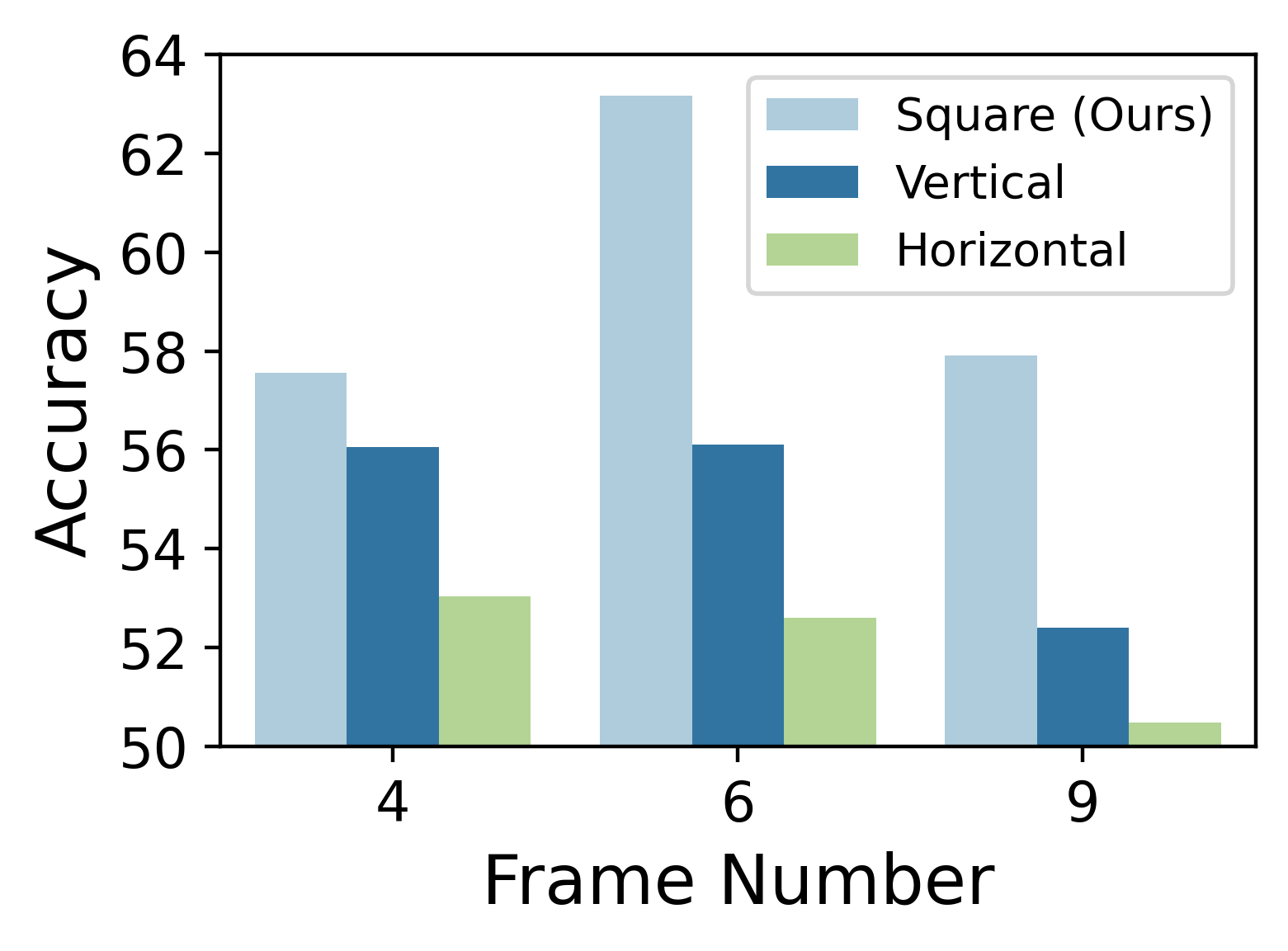}
        \caption{NExT-QA}
        \label{fig:analysis_shape_and_ordering:shape2}
    \end{subfigure}
    \caption{Analysis on the shape of image grid. LLaVA v1.6 with 7B parameters was used.
    }
    \vspace{-4pt}
    \label{fig:analysis_shape}
\end{figure}

For packing $N$ sampled frames, i.e., images, into an image grid, we consider three options for arranging them. The first is to place them from top left to bottom right such that the grid can have a shape of square. When $\sqrt{N}$ is an integer, a perfect square can be formed. For instance, $3\times3$ grid can be formed when $N=9$. When $N$ can be expressed as $N=M(M-1)$ for an integer $M$, we choose $M\times (M-1)$ as the arrangement such that the shape is close to a square. The other two options are placing all sampled images in a vertical column~($N\times 1$) or in a horizontal row~($1\times N$). The experiment results for two datasets are shown in  \Cref{fig:analysis_shape_and_ordering:shape1} and  \Cref{fig:analysis_shape_and_ordering:shape2}. For all six combinations that we have explored, square shape outperforms the other two options. This may be attributed to the likelihood that images utilized in training VLMs typically have an aspect ratio close to one.

\begin{figure}[]
    \centering
    \begin{subfigure}{0.40\linewidth}
        \includegraphics[width=\linewidth]{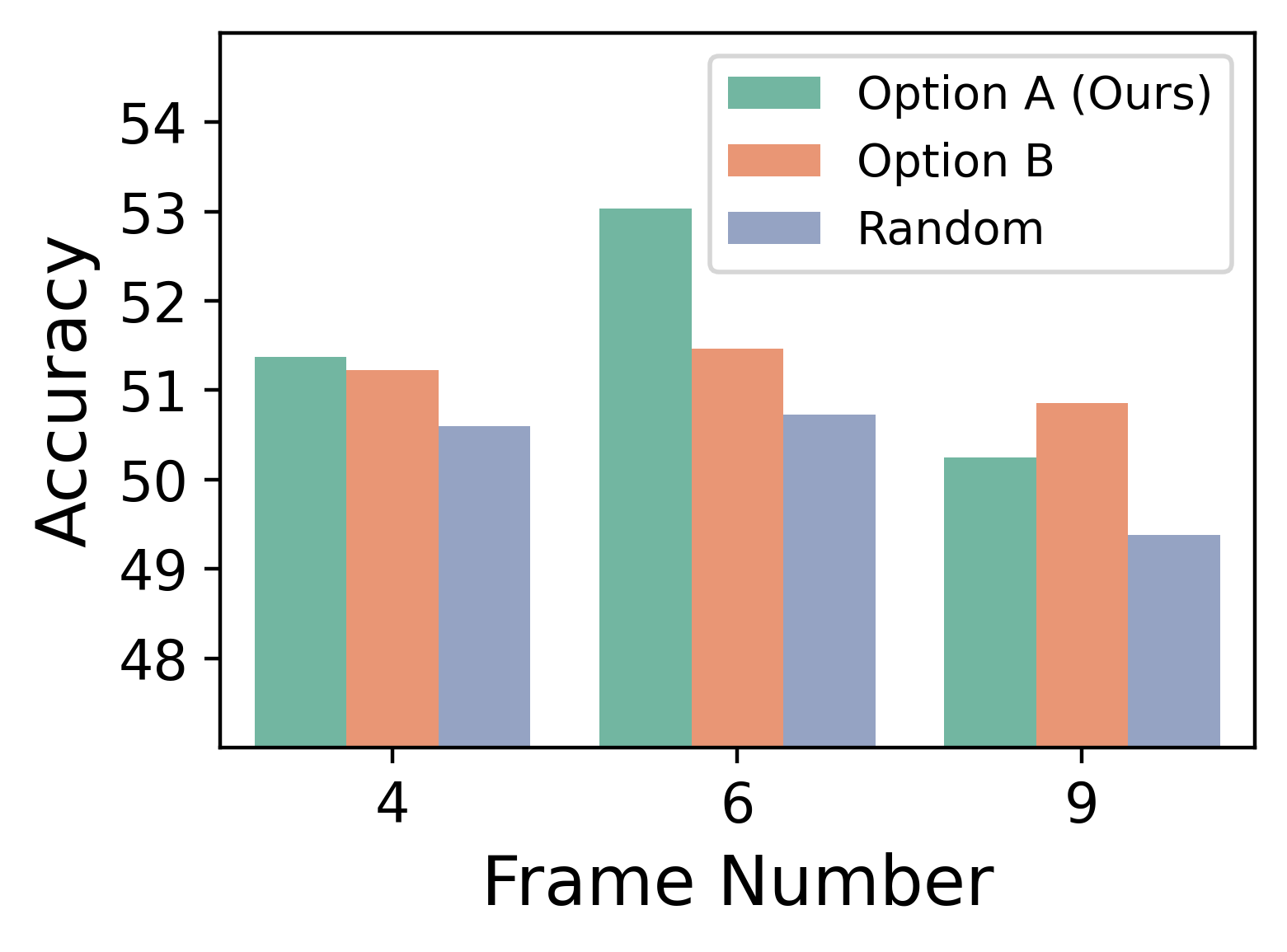}
        \caption{ActivityNet-QA}
        \label{fig:analysis_shape_and_ordering:ordering1}
    \end{subfigure}
    \begin{subfigure}{0.40\linewidth}
        \includegraphics[width=\linewidth]{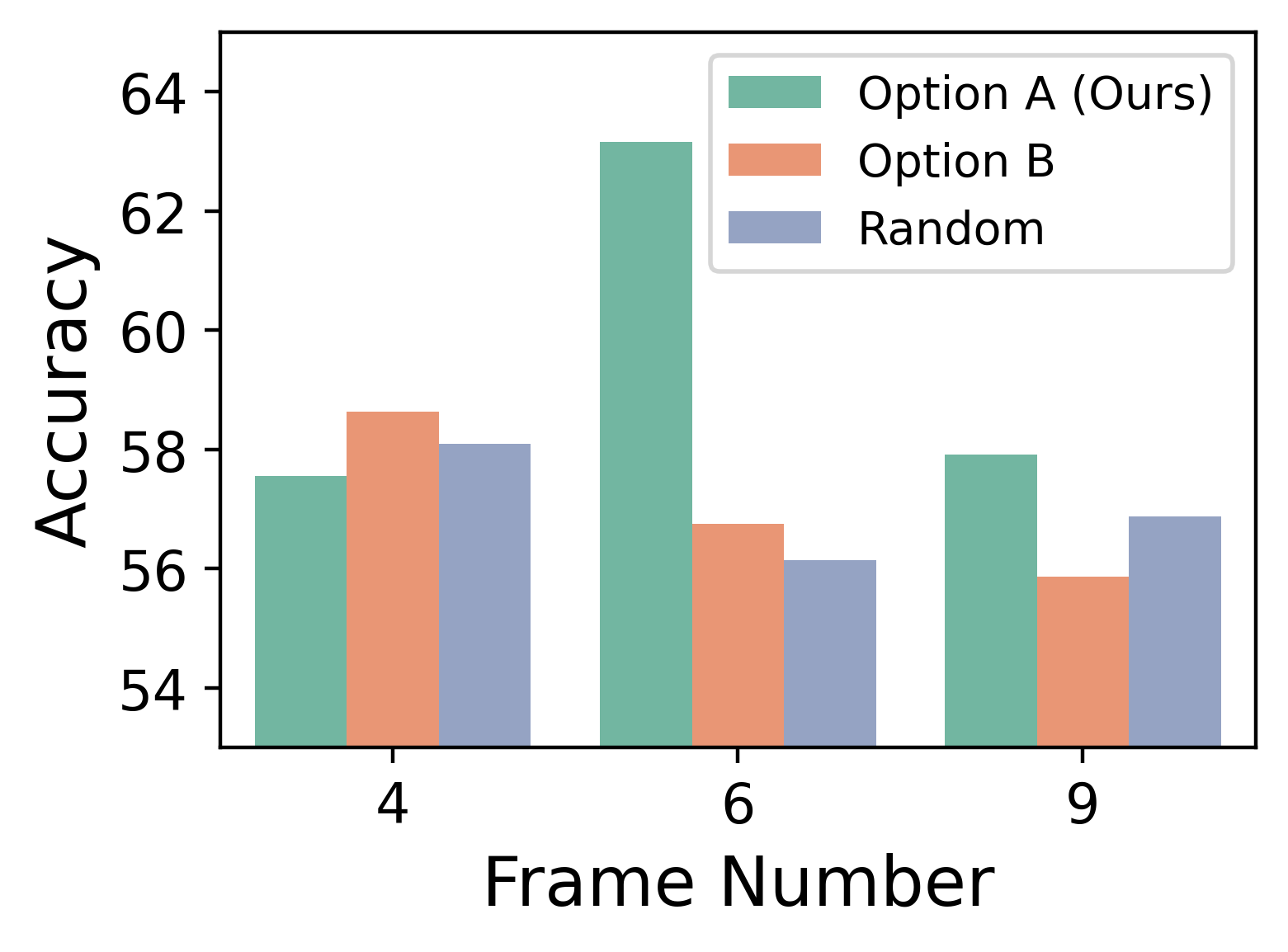}
        \caption{NExT-QA}
        \label{fig:analysis_shape_and_ordering:ordering2}
    \end{subfigure}
    \caption{Analysis on the ordering of the sampled frames. LLaVA v1.6 with 7B parameters was used.
    }
    \vspace{-10pt}
    \label{fig:analysis_ordering}
\end{figure}

\begin{figure}
    \begin{center}
    \begin{subfigure}{0.45\linewidth}
        \includegraphics[width=\linewidth]{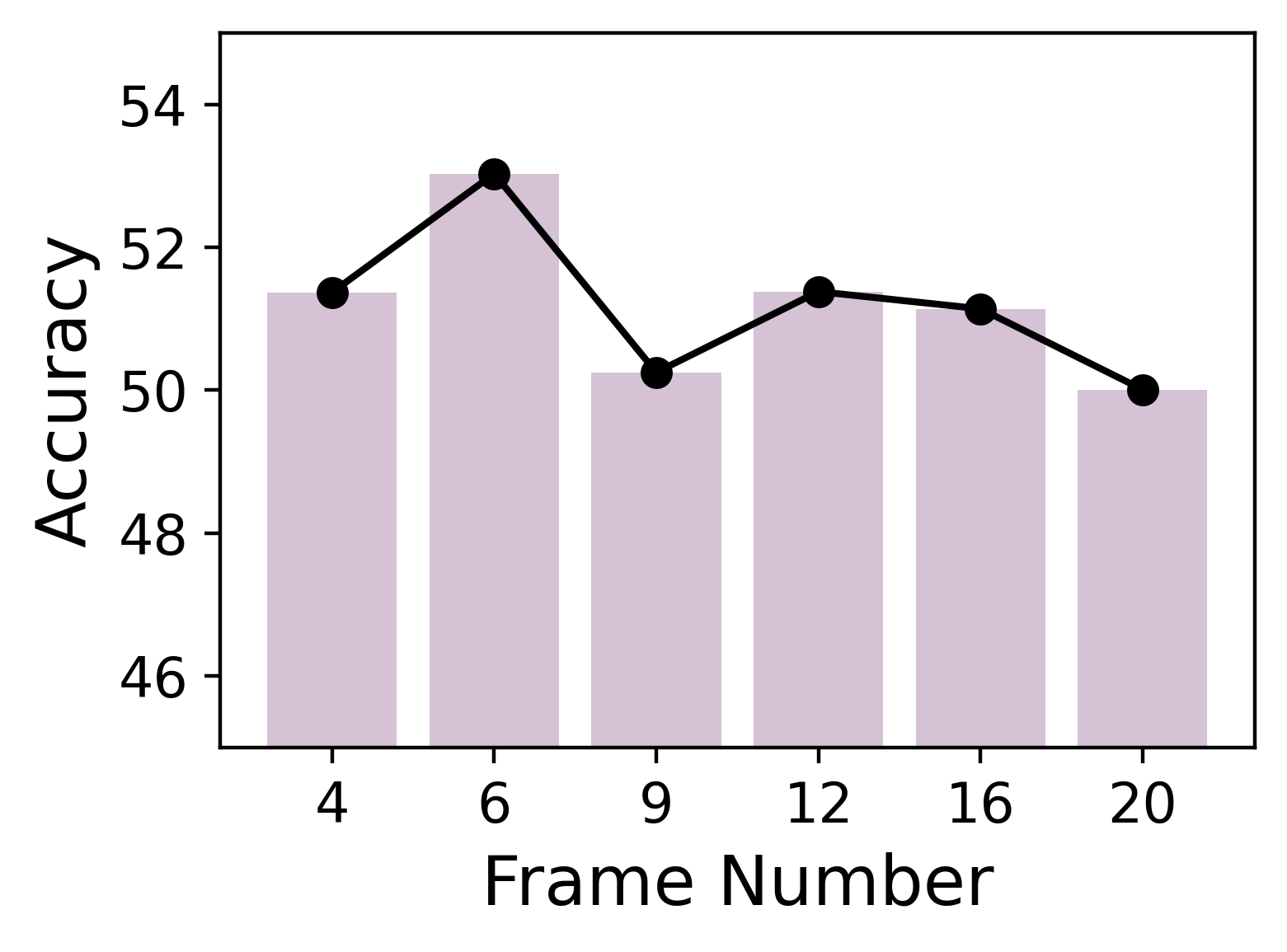}
        \caption{ActivityNet-QA}
    \end{subfigure}
    \begin{subfigure}{0.45\linewidth}
        \includegraphics[width=\linewidth]{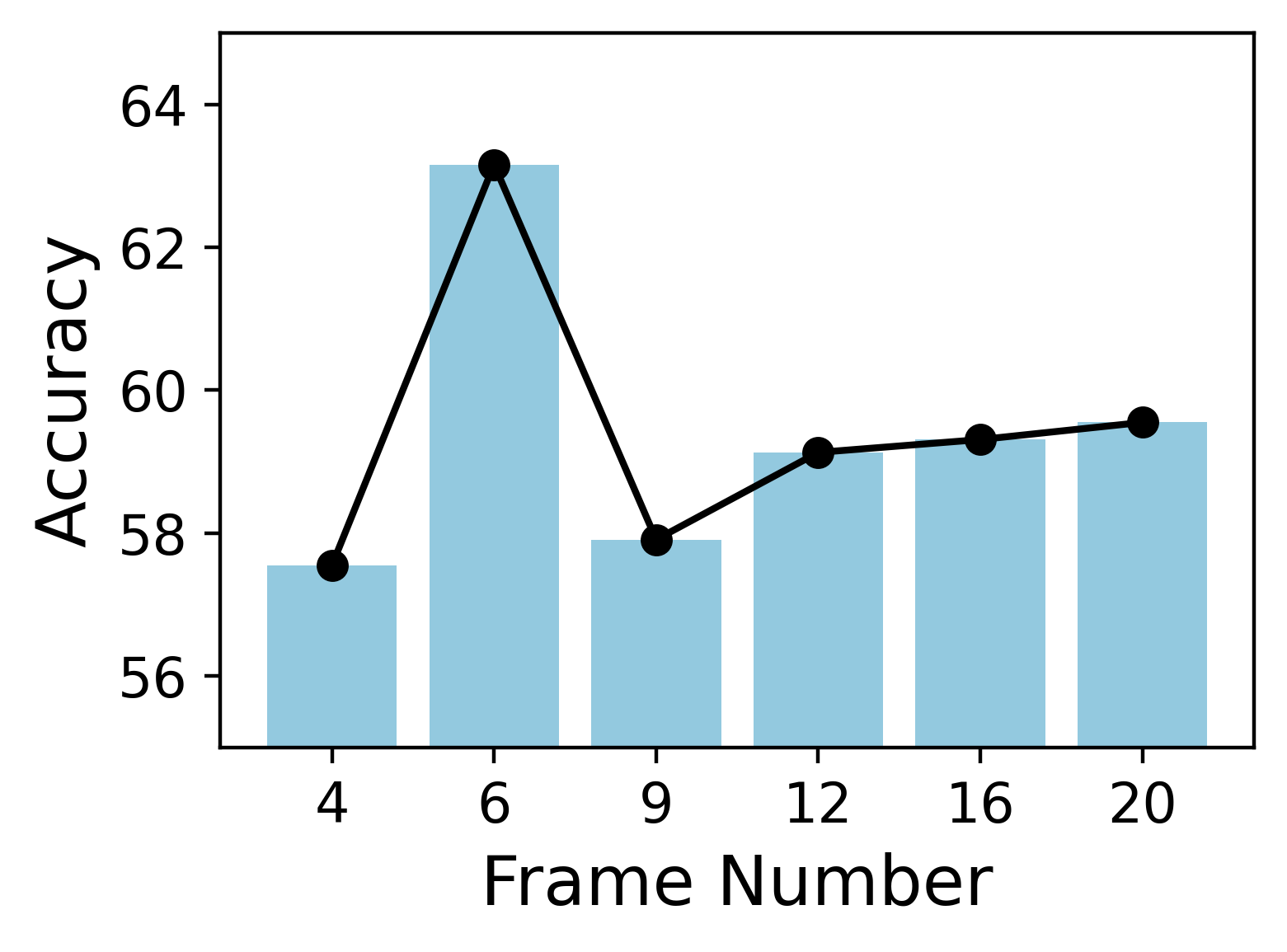}
        \caption{NExT-QA}
    \end{subfigure}
    \end{center}
    \caption{Analysis on the number of frames to include in an image grid.}
    \label{fig:analysis_number_of_frames}
\end{figure}

\subsubsection{Ordering of the sampled frames:} 
\label{subsec:order_of_sampled_frames}
For a square-shaped grid, we have always arranged images from top left to bottom right. In this case, the ordering can be done in two different ways. The first is to sweep the first row first from left to right and to continue to the next row (Option A) and the second is to sweep the first column first from top to bottom and to continue to the next column (Option B). We have compared the two, additionally with the baseline of random ordering, and the experiment results for two datasets are shown in  \Cref{fig:analysis_shape_and_ordering:ordering1} and  \Cref{fig:analysis_shape_and_ordering:ordering2}. While both options outperform the random ordering, Option A performs better than Option B on the average.

\subsubsection{Number of sampled frames to include:} 
\label{subsec:number_of_sampled_frames}
Given the decision to utilize a square-shaped grid, we selected $N$ that either satisfies `$\sqrt{N}$ is an integer' or `$N$ can be expressed as $M(M-1)$ for an integer $M$'. The outcomes of the investigation are shown in \cref{fig:analysis_number_of_frames}, demonstrating that a grid size of $N=6$ yields the best performance in both experiments conducted. A notable trade-off exists between the resolution of individual sampled frames and the aggregate number of frames that can be accommodated within an image grid. This trade-off is discussed further in Section~\ref{subsec:resolution}.

\subsection{Ablation study -- single frame vs. image grid}
\label{subsec:ablation_single_image_vs_image_grid}
An ablation study was performed to evaluate the advantage of utilizing an image grid compared to the baseline approach of using a single random frame. The results are presented in \cref{tab:single_image_vs_image_grid_1} for two multiple-choice benchmarks and in \cref{tab:single_image_vs_image_grid_2} for the text generation benchmark. The image grid approach consistently showed positive improvements in all tested scenarios. A notable increase in performance was observed with GPT-4V, as indicated in \cref{tab:single_image_vs_image_grid_1}. The degree of enhancement seems to be dependent on various factors, such as the choice of VLM. Nonetheless, the consistent performance gains across diverse conditions highlight the effectiveness of the image grid strategy.

\begin{table}[]
\centering
\caption{Single frame vs. image grid: ablation results for multiple-choice benchmarks. * denotes that we use 20\% of validation due to evaluation cost. }
\label{tab:single_image_vs_image_grid_1}
\scalebox{0.9}{
\begin{tabular}{c|c|c|c|cccc}
\toprule
\multirow{2}{*}{\textbf{Model}} & \multirow{2}{*}{\makecell{\textbf{LLM}\\\textbf{Size}}} & \multirow{2}{*}{\makecell{\textbf{Input Image}\\\textbf{Type}}} & \multicolumn{1}{c|}{\textbf{ActivityNet-QA}}                            & \multicolumn{4}{c}{\textbf{NExT-QA}}                                                                              \\
                                &                                    &                                            & \textbf{Accuracy}  & \textbf{Cas.} & \textbf{Tem.} & \textbf{Des.} & \textbf{Avg. Acc.} \\ \midrule
\multirow{2}{*}{LLaVA v1.6}     & \multirow{2}{*}{7B}                & Single Frame                               & 49.8                               & 59.1                          & 51.7                          & 71.6                          & 58.6         \\
                                &                                    & Image Grid (6)                             & 53.0 (\textbf{+3.2})                              & 63.1                           & 57.3                           & 74.9                           & 63.1 (\textbf{+4.5})          \\ \midrule
\multirow{2}{*}{LLaVA v1.6}     & \multirow{2}{*}{13B}               & Single Frame                               & 49.0                              & 58.7                          & 50.2                          & 70.1                           & 57.8         \\
                                &                                    & Image Grid (6)                             & 54.2 (\textbf{+5.2})                              & 61.6                           & 55.7                           & 70.8                           & 61.2 (\textbf{+3.4})          \\ \midrule
\multirow{2}{*}{LLaVA v1.6}     & \multirow{2}{*}{34B}               & Single Frame                               & 51.3                              & 68.2                           & 60.9                           & 76.8                           & 68.1          \\
                                &                                    & Image Grid (6)                             & 57.8 (\textbf{+6.5})                              & 72.2                          & 65.7                          & 77.3                          & 70.9 (\textbf{+2.8})          \\ \midrule
\multirow{2}{*}{GPT-4V*}   & \multirow{2}{*}{GPT-4}           & Single frame                               &         26.3                & 51.5                              & 34.2                              & 63.4         & 47.8     \\
                                &                                    & Image Grid (6)                             & 54.7 (\textbf{+28.4})                               & 70.0                           & 63.9                           & 75.1                           & 68.8 (\textbf{+21.0})         \\ \bottomrule
\end{tabular} 
}
\end{table}

\begin{table}[]
\centering
\caption{Single frame vs. image grid: ablation results for text generation benchmark.}
\label{tab:single_image_vs_image_grid_2}
\scalebox{0.9}{
\begin{tabular}{c|c|c|cccccc}
\toprule
\textbf{Model}              & \makecell{\textbf{LLM}\\\textbf{Size}}        & \makecell{\textbf{Input Image}\\\textbf{Type}} & \textbf{CI} & \textbf{DO} & \textbf{CU} & \textbf{TU} & \textbf{CO} & \textbf{Avg. Score} \\ \midrule
\multirow{2}{*}{CogAgent}   & \multirow{2}{*}{7B}      & Single Frame              & 3.13        & 2.59        & 3.41        & 2.29        & 2.83        & 2.85                \\
                            &                          & Image Grid (6)            & 3.26        & 2.76        & 3.57        & 2.34        & 3.28        & 3.04 \textbf{(+0.19)}                   \\ \midrule
\multirow{2}{*}{GPT-4V}       & \multirow{2}{*}{GPT-4} & Single Frame              & 2.67        & 2.18        & 2.86        & 2.67        & 2.39        & 2.55                \\
                            &                          & Image Grid (6)            & 3.40        & 2.80        & 3.61        & 2.89        & 3.13        & 3.17 \textbf{(+0.62)}                \\ \bottomrule
\end{tabular}
}
\end{table}

\subsection{Ablation study -- prompt design}
An additional ablation study was undertaken to determine the effect of prompt design on model performance. In the baseline condition, the VLM received only the question along with the image grid, without any additional context or instructions. As an enhanced baseline, we included details of the grid's structure, specifying how the sampled frames were arranged within the image grid. These two baselines were then evaluated against our refined prompt design, depicted in \cref{fig:videogrid_overview}, which incorporated supplementary guidance on video question answering. The results are presented in \cref{tab:ablation_prompt_design1} for three open-ended VQA benchmarks and in \cref{tab:ablation_prompt_design2} for the text generation benchmark.
Initially, we noted performance enhancements in the ActivityNet benchmark upon adding grid guidance prompt, a critical element for tasks requiring reasoning to understand temporal sequence across long video sequences. Furthermore, larger models demonstrated more considerable gains when the prompt explicitly described the image grid configuration. Including reasoning guidance in the prompt yielded a universal boost in performance, providing the model with explicit strategies for tackling tasks related to the image grid.
We simplified our prompt design for multiple-choice benchmarks to include only the question and the grid structure information, omitting additional reasoning guidance. This approach was influenced by the structured nature of multiple-choice queries, in which the provided answer 

\FloatBarrier
\begin{table}[]
\centering
\caption{Ablation study of three prompt designs across three open-ended VQA benchmarks. * denotes that we use 20\% of validation due to evaluation cost.}
\label{tab:ablation_prompt_design1}
\scalebox{0.7}{
\begin{tabular}{c|l|cc|cc|cc}
\toprule
\multirow{2}{*}{\textbf{VLM}} & \multicolumn{1}{c|}{\multirow{2}{*}{~~~~~~~~~~\textbf{Prompt}~~~~~~~~~~}} & \multicolumn{2}{c|}{\textbf{ActivityNet-QA}} & \multicolumn{2}{c|}{\textbf{MSVD-QA}} & \multicolumn{2}{c}{\textbf{MSRVTT-QA}}  \\ 
 & \multicolumn{1}{c|}{} & \textbf{~~Acc.~~} & \textbf{Score}  & \textbf{Acc.} & \textbf{Score} & \textbf{~Acc.~} & \textbf{Score} \\
\midrule
\multirow{3}{*}{LLaVA v1.6 7B} & ~Question only & 51.5 & 3.38 & 73.9 & 3.96 & 56.8 & 3.34 \\
& ~+ Grid guidance & 53.0 & 3.42 & 72.3 & 3.90 & 56.4 & 3.29  \\
& ~+ Reasoning guidance & \textbf{54.3} & 3.36 & \textbf{78.8} & 4.10 & \textbf{63.7} & 3.53 \\
\midrule
\multirow{3}{*}{LLaVA v1.6 13B} & ~Question only & 29.6 & 3.62 & 74.5 & 4.01 & 62.6 & 3.43 \\
& ~+ Grid guidance & 54.2 & 3.53 & 72.9 & 4.04 & 56.5 & 3.43 \\
& ~+ Reasoning guidance & \textbf{57.1} & 3.53 & \textbf{77.4} & 4.09 & \textbf{62.6} & 3.43 \\
\midrule
\multirow{3}{*}{LLaVA v1.6 34B} & ~Question only & 56.7 & 3.52 & 75.4 & 3.94 & 51.6 & 3.11 \\
& ~+ Grid guidance & 57.8 & 3.66 & 76.1 & 4.06 & 58.8 & 3.46 \\
& ~+ Reasoning guidance & \textbf{58.4} & 3.50 & \textbf{79.6} & 4.08 & \textbf{62.4} & 3.45 \\
\midrule
\multirow{3}{*}{GPT-4V*} & ~Question only & 49.1 & 3.18 & 68.7 & 3.60 & 51.7 & 3.02 \\
& ~+ Grid guidance & 53.4 & 3.41 & 77.0 & 3.97 & 59.6 & 3.39 \\
& ~+ Reasoning guidance & \textbf{54.7} & 3.45 & \textbf{77.6} & 3.97 & \textbf{64.3} & 3.47 \\
\bottomrule
\end{tabular}
}
\end{table}

\noindent options naturally allow VLM to utilize an image grid to select the best answer to the question. Given the proficiency of VLMs in handling multiple-choice questions without supplementary instructions, we opted not to incorporate additional guidance into our prompt design for these benchmarks.

\section{Discussion}
\subsection{Size of LLM and VQA performance}
\label{subsec:diss_LLM_size}
In general, larger LLMs are known to perform better. 
However, the same claim cannot be made for the VQA experiment results of our IG-VLM. Experiments involving LLaVA v1.6 within IG-VLM have indeed indicated a potential improvement in performance with scaling of LLM sizes; however, this pattern does not uniformly apply across various open-ended and multiple-choice VQA benchmarks. Interestingly, despite the significant size difference between GPT-4V and LLaVA v1.6 34B, the latter demonstrated superior performance in three out of five open-ended VQA. Similarly, in the case of multiple-choice VQA, LLaVA v1.6 34B excelled in three out of five instances.

\subsection{Spatial-temporal trade-offs of IG-VLM}
\label{subsec:resolution}

The image grid utilized in IG-VLM is relatively high-resolution, which may result in a decreased resolution and altered aspect ratios during the preprocessing of the image grid in VLM. Additionally, the image grid might be partially cropped, potentially causing modifications or deformations of the sampled frames. This can lead to a loss of spatial detail in the sampled frames represented on the image grid. This negative impact might apply not only to spatial information but also to temporal information. The VLM understands each sampled frame individually and comprehends the video based on the sequence of sampled frames. 

\label{sec:prompt_design_ablation}
\begin{table}[]
\centering
\caption{Ablation study of three prompt designs for the text generation benchmark.} 

\label{tab:ablation_prompt_design2}
\scalebox{0.8}{
\begin{tabular}{c|l|cccccc}
\toprule
\textbf{VLM} & \multicolumn{1}{c|}{~~~~~~~~~~~\textbf{Prompt}~~~~~~~~~~~} & ~\textbf{CI} & \textbf{DO} & \textbf{CU} & \textbf{TU} & \textbf{CO} & \makecell{\textbf{Avg.}\\\textbf{Score}} \\
\midrule
\multirow{3}{*}{LLaVA v1.6 7B} & ~Question only & ~2.64 & 2.58 & 3.04 & 2.08 & 3.08 & 2.68 \\
& ~+ Grid guidance & ~2.53 & 2.49 & 2.89 & 2.06 & 3.02 & 2.60 \\
& ~+ Reasoning guidance & ~3.11 & 2.78 & 3.51 & 2.44 & 3.29 & \textbf{3.03} \\
\midrule
\multirow{3}{*}{LLaVA v1.6 13B} & ~Question only & ~2.66 & 2.60 & 3.04 & 2.16 & 3.03 & 2.72 \\
& ~+ Grid guidance & ~2.51 & 2.54 & 2.83 & 2.17 & 3.03 & 2.62 \\
& ~+ Reasoning guidance & ~3.17 & 2.79 & 3.52 & 2.51 & 3.25 & \textbf{3.05} \\
\midrule
\multirow{3}{*}{LLaVA v1.6 34B} & ~Question only & ~2.86 & 2.69 & 3.18 & 2.12 & 3.38 & 2.85 \\
& ~+ Grid guidance & ~3.21 & 2.68 & 3.19 & 2.31 & 3.01 & 2.88 \\
& ~+ Reasoning guidance & ~3.21 & 2.87 & 3.54 & 2.51 & 3.34 & \textbf{3.09} \\
\midrule
\multirow{3}{*}{GPT-4V} & ~Question only & ~2.86 & 2.48 & 2.85 & 2.96 & 3.13 & 2.86 \\
& ~+ Grid guidance & ~3.08 & 2.91 & 3.31 & 2.89 & 3.23 & 3.08 \\
& ~+ Reasoning guidance & ~3.40 & 2.80 & 3.61 & 2.89 & 3.13 & \textbf{3.17} \\
\bottomrule
\end{tabular}
}
\end{table}

However, due to the loss of spatial information mentioned earlier, there could be insufficient understanding or information about each sampled frame, which in turn, degrades the analysis of temporal aspects. This indicates that the number of frames represented in the image grid, the shape of the image grid, as well as the resolution of the vision encoder of the VLM, can all influence the performance of IG-VLM.

\subsection{Comparing Single-Step vs. Multi-Step Prompting}

IG-VLM aims to excel in VQA through a simple prompt that encompasses grid guidance and reasoning guidance. The utilization of a straightforward prompt has yielded favorable results across various benchmarks. Given that our prompt facilitates a single-step answer generation process, we conducted an analysis to explore the potential of multi-step reasoning, encouraging a step-by-step intermediate rationale for LLM.
To enable the incorporation of multi-step reasoning, we applied the Chain-of-Thought (CoT) methodology~\cite{kojima2022large,wang2023selfconsistency,Wang2023PlanandSolvePI}, a technique used to enhance the reasoning capabilities of LLM.
In order to fairly compare single-step and multi-step reasoning, grid guidance was applied across all experiment steps when exploring the potential benefits of multi-step reasoning.
Experimental results revealed some cases where prompts designed for multi-step reasoning led to improvements in accuracy. The corresponding experimental results can be found in Appendix \ref{appendix:CoT_prompt_result}. Considering that IG-VLM combines grid guidance with reasoning guidance, there exists a potential for enhancing reasoning in VLM through the integration of these elements of multi-step reasoning.
The prompt template used in this experiment can be found in Appendix \ref{appendix:CoT_prompt_template}.

\section{Conclusions}
In this work, we propose the Image Grid Vision Language Model~(IG-VLM), a novel approach that uses an image grid to transmit multiple video frames to a Vision Language Model~(VLM). The image grid can convey both spatial and temporal information within a single image format, a capability validated through extensive experiments across ten video question answering benchmarks. However, a notable drawback of the image grid is its inherent limitation of discarding most of the video frames. Addressing this limitation remains as an area for future exploration.

\section{Acknowledgements}
This work was supported by NRF(No. NRF-2020R1A2C2007139) and IITP([NO.2021-0-01343, Artificial Intelligence Graduate School Program (Seoul National University)], [No. RS-2023-00235293])

\clearpage

\bibliographystyle{splncs04}
\bibliography{egbib}

\begin{thebibliography}{10}
\providecommand{\url}[1]{\texttt{#1}}
\providecommand{\urlprefix}{URL }
\providecommand{\doi}[1]{https://doi.org/#1}

\bibitem{achiam2023gpt}
Achiam, J., Adler, S., Agarwal, S., Ahmad, L., Akkaya, I., Aleman, F.L., Almeida, D., Altenschmidt, J., Altman, S., Anadkat, S., et~al.: Gpt-4 technical report. arXiv preprint arXiv:2303.08774  (2023)

\bibitem{Koyejo2022FLAMINGONIPS}
Alayrac, J.B., Donahue, J., Luc, P., Miech, A., Barr, I., Hasson, Y., Lenc, K., Mensch, A., Millican, K., Reynolds, M., Ring, R., Rutherford, E., Cabi, S., Han, T., Gong, Z., Samangooei, S., Monteiro, M., Menick, J.L., Borgeaud, S., Brock, A., Nematzadeh, A., Sharifzadeh, S., Bi\'{n}kowski, M.a., Barreira, R., Vinyals, O., Zisserman, A., Simonyan, K.: Flamingo: a visual language model for few-shot learning. In: Koyejo, S., Mohamed, S., Agarwal, A., Belgrave, D., Cho, K., Oh, A. (eds.) Adv. Neural Inform. Process. Syst. vol.~35, pp. 23716--23736. Curran Associates, Inc. (2022)

\bibitem{Anil2023PaLM2T}
Anil, R., Dai, A.M., Firat, O., Johnson, M., Lepikhin, D., Passos, A., Shakeri, S., Taropa, E., Bailey, P., Chen, Z., et~al.: Palm 2 technical report. arXiv preprint arXiv:2305.10403  (2023)

\bibitem{Brown2020LanguageMA}
Brown, T., Mann, B., Ryder, N., Subbiah, M., Kaplan, J.D., Dhariwal, P., Neelakantan, A., Shyam, P., Sastry, G., Askell, A., et~al.: Language models are few-shot learners. Advances in neural information processing systems  \textbf{33},  1877--1901 (2020)

\bibitem{chen2021evaluating}
Chen, M., Tworek, J., Jun, H., Yuan, Q., Pinto, H.P.d.O., Kaplan, J., Edwards, H., Burda, Y., Joseph, N., Brockman, G., et~al.: Evaluating large language models trained on code. arXiv preprint arXiv:2107.03374  (2021)

\bibitem{Choudhury2023ZeroShotVQ}
Choudhury, R., Niinuma, K., Kitani, K.M., Jeni, L.A.: Zero-shot video question answering with procedural programs. ArXiv abs/2312.00937  (2023)

\bibitem{Chowdhery2022PaLMSL}
Chowdhery, A., Narang, S., Devlin, J., Bosma, M., Mishra, G., Roberts, A., Barham, P., Chung, H.W., Sutton, C., et~al, S.G.: Palm: Scaling language modeling with pathways. J. Mach. Learn. Res.  \textbf{24},  240:1--240:113 (2022)

\bibitem{Qwen-Audio}
Chu, Y., Xu, J., Zhou, X., Yang, Q., Zhang, S., Yan, Z., Zhou, C., Zhou, J.: Qwen-audio: Advancing universal audio understanding via unified large-scale audio-language models. arXiv preprint arXiv:2311.07919  (2023)

\bibitem{Dai2023InstructBLIPTG}
Dai, W., Li, J., Li, D., Tiong, A.M.H., Zhao, J., Wang, W., Li, B.A., Fung, P., Hoi, S.C.H.: Instructblip: Towards general-purpose vision-language models with instruction tuning. ArXiv  \textbf{abs/2305.06500} (2023)

\bibitem{deshmukh2023pengi}
Deshmukh, S., Elizalde, B., Singh, R., Wang, H.: Pengi: An audio language model for audio tasks. In: Adv. Neural Inform. Process. Syst. (2023)

\bibitem{Grauman2021Ego4DAT}
Grauman, K., Westbury, A., Byrne, E., Chavis, Z., Furnari, A., Girdhar, R., Hamburger, J., Jiang, H., Liu, M., et~al, X.L.: Ego4d: Around the world in 3,000 hours of egocentric video. IEEE Conf. Comput. Vis. Pattern Recog. pp. 18995--19012 (2022)

\bibitem{Guo2022FromIT}
Guo, J., Li, J., Li, D., Tiong, A.M.H., Li, B., Tao, D., Hoi, S.: From images to textual prompts: Zero-shot visual question answering with frozen large language models. In: Proceedings of the IEEE/CVF Conference on Computer Vision and Pattern Recognition. pp. 10867--10877 (2023)

\bibitem{hong2023cogagent}
Hong, W., Wang, W., Lv, Q., Xu, J., Yu, W., Ji, J., Wang, Y., Wang, Z., Zhang, Y., Li, J., Xu, B., Dong, Y., Ding, M., Tang, J.: Cogagent: A visual language model for gui agents. ArXiv  \textbf{abs/2312.08914} (2023)

\bibitem{huang2023audiogpt}
Huang, R., Li, M., Yang, D., Shi, J., Chang, X., Ye, Z., Wu, Y., Hong, Z., Huang, J., Liu, J., et~al.: Audiogpt: Understanding and generating speech, music, sound, and talking head. arXiv preprint arXiv:2304.12995  (2023)

\bibitem{Jin2023ChatUniViUV}
Jin, P., Takanobu, R., Zhang, C., Cao, X., Yuan, L.: Chat-univi: Unified visual representation empowers large language models with image and video understanding  (2024)

\bibitem{kojima2022large}
Kojima, T., Gu, S.S., Reid, M., Matsuo, Y., Iwasawa, Y.: Large language models are zero-shot reasoners. Adv. Neural Inform. Process. Syst.  \textbf{35},  22199--22213 (2022)

\bibitem{lei2018tvqa}
Lei, J., Yu, L., Bansal, M., Berg, T.L.: Tvqa: Localized, compositional video question answering. In: Conf. Empirical Methods in Natural Language Processing (2018)

\bibitem{Li2023IntentQACV}
Li, J., Wei, P., Han, W., Fan, L.: Intentqa: Context-aware video intent reasoning. In: Int. Conf. Comput. Vis. pp. 11963--11974 (2023)

\bibitem{Li2023BLIP2BL}
Li, J., Li, D., Savarese, S., Hoi, S.C.H.: Blip-2: Bootstrapping language-image pre-training with frozen image encoders and large language models. In: Porc. Int. Conf. Machine Learn. (2023)

\bibitem{Li2022BLIPBL}
Li, J., Li, D., Xiong, C., Hoi, S.C.H.: Blip: Bootstrapping language-image pre-training for unified vision-language understanding and generation. In: Porc. Int. Conf. Machine Learn. (2022)

\bibitem{Li2023VideoChatCV}
Li, K., He, Y., Wang, Y., Li, Y., Wang, W., Luo, P., Wang, Y., Wang, L., Qiao, Y.: Videochat: Chat-centric video understanding. ArXiv abs/2305.06355  (2023)

\bibitem{li2023mvbench}
Li, K., Wang, Y., He, Y., Li, Y., Wang, Y., Liu, Y., Wang, Z., Xu, J., Chen, G., Luo, P., et~al.: Mvbench: A comprehensive multi-modal video understanding benchmark  (2023)

\bibitem{Li2023LLaMAVIDAI}
Li, Y., Wang, C., Jia, J.: Llama-vid: An image is worth 2 tokens in large language models. ArXiv abs/2311.17043  (2023)

\bibitem{Jang2017TGIFQATS}
Li, Y., Song, Y., Cao, L., Tetreault, J., Goldberg, L., Jaimes, A., Luo, J.: Tgif: A new dataset and benchmark on animated gif description. In: Proceedings of the IEEE Conference on Computer Vision and Pattern Recognition. pp. 4641--4650 (2016)

\bibitem{ModalityGap}
Liang, V.W., Zhang, Y., Kwon, Y., Yeung, S., Zou, J.Y.: Mind the gap: Understanding the modality gap in multi-modal contrastive representation learning. Adv. Neural Inform. Process. Syst.  \textbf{35},  17612--17625 (2022)

\bibitem{Lin2023VideoLLaVALU}
Lin, B., Zhu, B., Ye, Y., Ning, M., Jin, P., Yuan, L.: Video-llava: Learning united visual representation by alignment before projection. ArXiv abs/2311.10122  (2023)

\bibitem{liu2024llavanext}
Liu, H., Li, C., Li, Y., Li, B., Zhang, Y., Shen, S., Lee, Y.J.: Llava-next: Improved reasoning, ocr, and world knowledge (January 2024)

\bibitem{liu2023llava}
Liu, H., Li, C., Wu, Q., Lee, Y.J.: Visual instruction tuning. In: NeurIPS (2023)

\bibitem{Luo2023ValleyVA}
Luo, R., Zhao, Z., Yang, M., Dong, J., Qiu, M.H., Lu, P., Wang, T., Wei, Z.: Valley: Video assistant with large language model enhanced ability. ArXiv abs/2306.07207  (2023)

\bibitem{Ma2023VistaLLaMARV}
Ma, F., Jin, X., Wang, H., Xian, Y., Feng, J., Yang, Y.: Vista-llama: Reliable video narrator via equal distance to visual tokens. ArXiv abs/2312.08870  (2023)

\bibitem{Maaz2023VideoChatGPTTD}
Maaz, M., Rasheed, H.A., Khan, S., Khan, F.S.: Video-chatgpt: Towards detailed video understanding via large vision and language models. ArXiv abs/2306.05424  (2023)

\bibitem{mangalam2023egoschema}
Mangalam, K., Akshulakov, R., Malik, J.: Egoschema: A diagnostic benchmark for very long-form video language understanding. In: Adv. Neural Inform. Process. Syst. (2024)

\bibitem{Shao_2023_CVPR}
Shao, Z., Yu, Z., Wang, M., Yu, J.: Prompting large language models with answer heuristics for knowledge-based visual question answering. In: IEEE Conf. Comput. Vis. Pattern Recog. pp. 14974--14983 (2023)

\bibitem{Song2023MovieChatFD}
Song, E., Chai, W., Wang, G., Zhang, Y., Zhou, H., Wu, F., Guo, X., Ye, T., Lu, Y., Hwang, J.N., Wang, G.: Moviechat: From dense token to sparse memory for long video understanding. ArXiv abs/2307.16449  (2023)

\bibitem{suris2023vipergpt}
Sur\'is, D., Menon, S., Vondrick, C.: Vipergpt: Visual inference via python execution for reasoning. Proceedings of IEEE International Conference on Computer Vision (ICCV)  (2023)

\bibitem{Tang2023SALMONNTG}
Tang, C., Yu, W., Sun, G., Chen, X., Tan, T., Li, W., Lu, L., Ma, Z., Zhang, C.: Salmonn: Towards generic hearing abilities for large language models. ArXiv abs/2310.13289  (2023)

\bibitem{Touvron2023LLaMAOA}
Touvron, H., Lavril, T., Izacard, G., Martinet, X., Lachaux, M.A., Lacroix, T., Rozi{\`e}re, B., Goyal, N., Hambro, E., Azhar, F., et~al.: Llama: Open and efficient foundation language models. arXiv preprint arXiv:2302.13971  (2023)

\bibitem{Wang2023PlanandSolvePI}
Wang, L., Xu, W., Lan, Y., Hu, Z., Lan, Y., Lee, R.K.W., Lim, E.P.: Plan-and-solve prompting: Improving zero-shot chain-of-thought reasoning by large language models. In: Annual Meeting of the Association for Computational Linguistics (2023)

\bibitem{Wang2023VLAPEV}
Wang, X., Liang, J., Wang, C.K., Deng, K., Lou, Y., Lin, M., Yang, S.: Vlap: Efficient video-language alignment via frame prompting and distilling for video question answering. ArXiv  \textbf{abs/2312.08367} (2023)

\bibitem{wang2023selfconsistency}
Wang, X., Wei, J., Schuurmans, D., Le, Q.V., Chi, E.H., Narang, S., Chowdhery, A., Zhou, D.: Self-consistency improves chain of thought reasoning in language models. In: Int. Conf. Learn. Represent. (2023)

\bibitem{Wang2022InternVideoGV}
Wang, Y., Li, K., Li, Y., He, Y., Huang, B., Zhao, Z., Zhang, H., Xu, J., Liu, Y., Wang, Z., Xing, S., Chen, G., Pan, J., Yu, J., Wang, Y., Wang, L., Qiao, Y.: Internvideo: General video foundation models via generative and discriminative learning. ArXiv abs/2212.03191  (2022)

\bibitem{wang2022simvlm}
Wang, Z., Yu, J., Yu, A.W., Dai, Z., Tsvetkov, Y., Cao, Y.: Sim{VLM}: Simple visual language model pretraining with weak supervision. In: Int. Conf. Learn. Represent. (2022)

\bibitem{Wang2023FillingTI}
Wang, Z., Chen, C., Li, P., Liu, Y.: Filling the image information gap for vqa: Prompting large language models to proactively ask questions. In: Conf. Empirical Methods in Natural Language Processing (2023)

\bibitem{wu2021star}
Wu, B., Yu, S., Chen, Z., Tenenbaum, J.B., Gan, C.: Star: A benchmark for situated reasoning in real-world videos. In: Thirty-fifth conference on neural information processing systems datasets and benchmarks track (Round 2) (2021)

\bibitem{xiao2021next}
Xiao, J., Shang, X., Yao, A., Chua, T.S.: Next-qa: Next phase of question-answering to explaining temporal actions. In: Proceedings of the IEEE/CVF conference on computer vision and pattern recognition. pp. 9777--9786 (2021)

\bibitem{xu2017video}
Xu, D., Zhao, Z., Xiao, J., Wu, F., Zhang, H., He, X., Zhuang, Y.: Video question answering via gradually refined attention over appearance and motion. In: Proceedings of the 25th ACM international conference on Multimedia. pp. 1645--1653 (2017)

\bibitem{Yang2022FrozenBiLM}
Yang, A., Miech, A., Sivic, J., Laptev, I., Schmid, C.: Zero-shot video question answering via frozen bidirectional language models. Adv. Neural Inform. Process. Syst.  \textbf{35},  124--141 (2022)

\bibitem{yang2024vidchapters}
Yang, A., Nagrani, A., Laptev, I., Sivic, J., Schmid, C.: Vidchapters-7m: Video chapters at scale. NeurIPS  \textbf{36} (2024)

\bibitem{yang2023vid2seq}
Yang, A., Nagrani, A., Seo, P.H., Miech, A., Pont-Tuset, J., Laptev, I., Sivic, J., Schmid, C.: Vid2seq: Large-scale pretraining of a visual language model for dense video captioning. In: IEEE Conf. Comput. Vis. Pattern Recog. pp. 10714--10726 (2023)

\bibitem{Ye2023mPLUGOwlME}
Ye, Q., Xu, H., Xu, G., Ye, J., Yan, M., Zhou, Y., Wang, J., Hu, A., Shi, P., Shi, Y., Li, C., Xu, Y., Chen, H., Tian, J., Qi, Q., Zhang, J., Huang, F.: mplug-owl: Modularization empowers large language models with multimodality. ArXiv  \textbf{abs/2304.14178} (2023)

\bibitem{Yu2022CoCaCC}
Yu, J., Wang, Z., Vasudevan, V., Yeung, L., Seyedhosseini, M., Wu, Y.: Coca: Contrastive captioners are image-text foundation models. In: Trans. Mach. Learn. Res. vol.~2022 (2022)

\bibitem{yu2023self}
Yu, S., Cho, J., Yadav, P., Bansal, M.: Self-chained image-language model for video localization and question answering. In: Adv. Neural Inform. Process. Syst. vol.~36 (2024)

\bibitem{yu2019activitynet}
Yu, Z., Xu, D., Yu, J., Yu, T., Zhao, Z., Zhuang, Y., Tao, D.: Activitynet-qa: A dataset for understanding complex web videos via question answering. In: AAAI. vol.~33, pp. 9127--9134 (2019)

\bibitem{Zhang2023ASL}
Zhang, C., Lu, T., Islam, M.M., Wang, Z., Yu, S., Bansal, M., Bertasius, G.: A simple llm framework for long-range video question-answering. ArXiv abs/2312.17235  (2023)

\bibitem{zhang2023VideoLLAMA}
Zhang, H., Li, X., Bing, L.: Video-{LL}a{MA}: An instruction-tuned audio-visual language model for video understanding. In: Conf. Empirical Methods in Natural Language Processing. pp. 543--553 (2023)

\bibitem{Zhang2023LLaMAAdapterEF}
Zhang, R., Han, J., Zhou, A., Hu, X., Yan, S., Lu, P., Li, H., Gao, P., Qiao, Y.: Llama-adapter: Efficient fine-tuning of language models with zero-init attention. arXiv preprint arXiv:2303.16199  (2023)

\end{thebibliography}
\newpage
\appendix
\section{Successful Case Examples of VQA by IG-VLM}
\label{appendix:example_IG-VLM}

\begin{figure}[]
  \centering
 \scalebox{0.895}{
  \includegraphics[width=\linewidth]{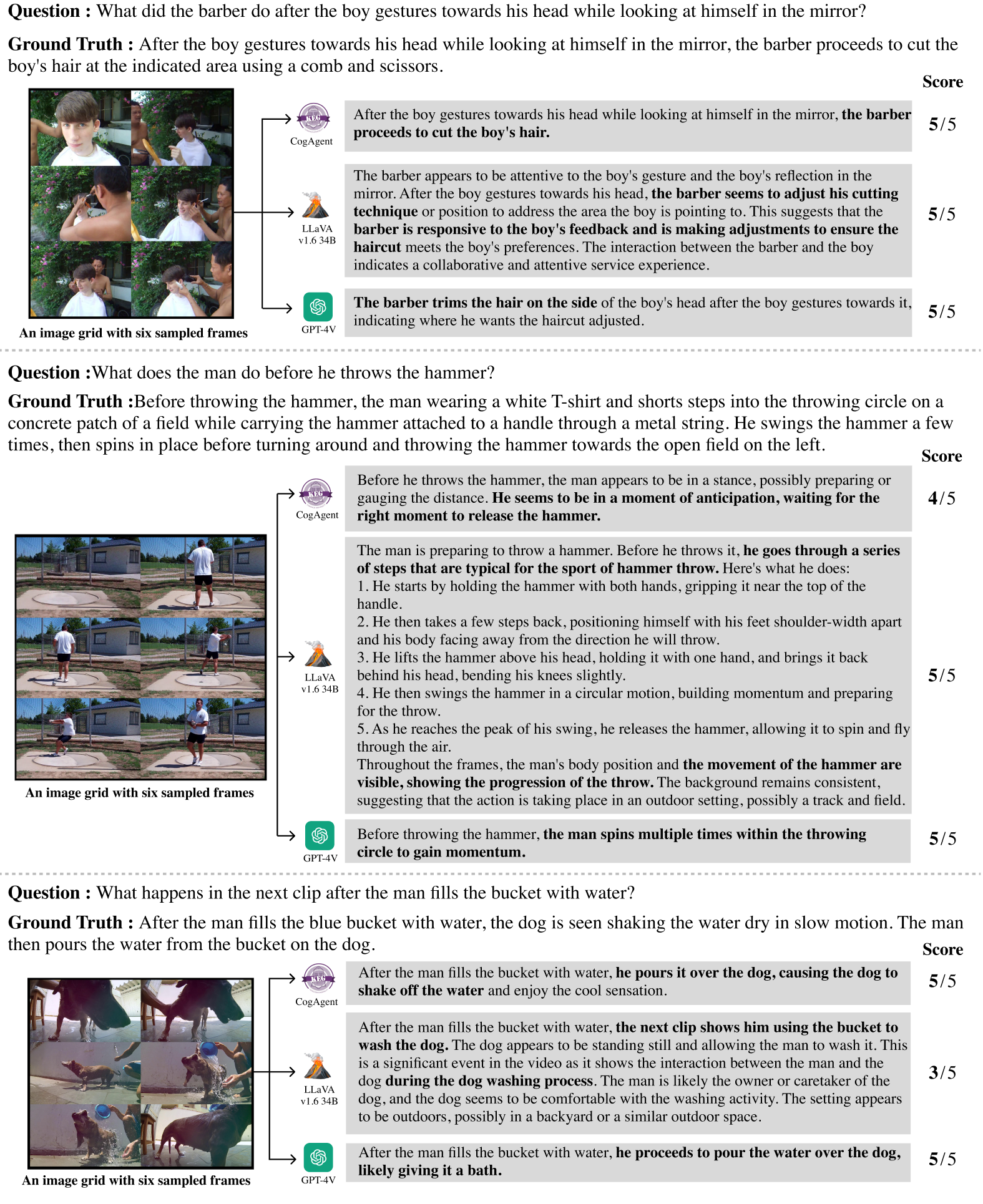}
}
  \caption{Successful case examples of zero-shot VQA on Text Generation Performance benchmark.}
  \label{fig:success_example_open1}
\end{figure}

\begin{figure}[]
  \centering
   \scalebox{0.93}{
  \includegraphics[width=\linewidth]{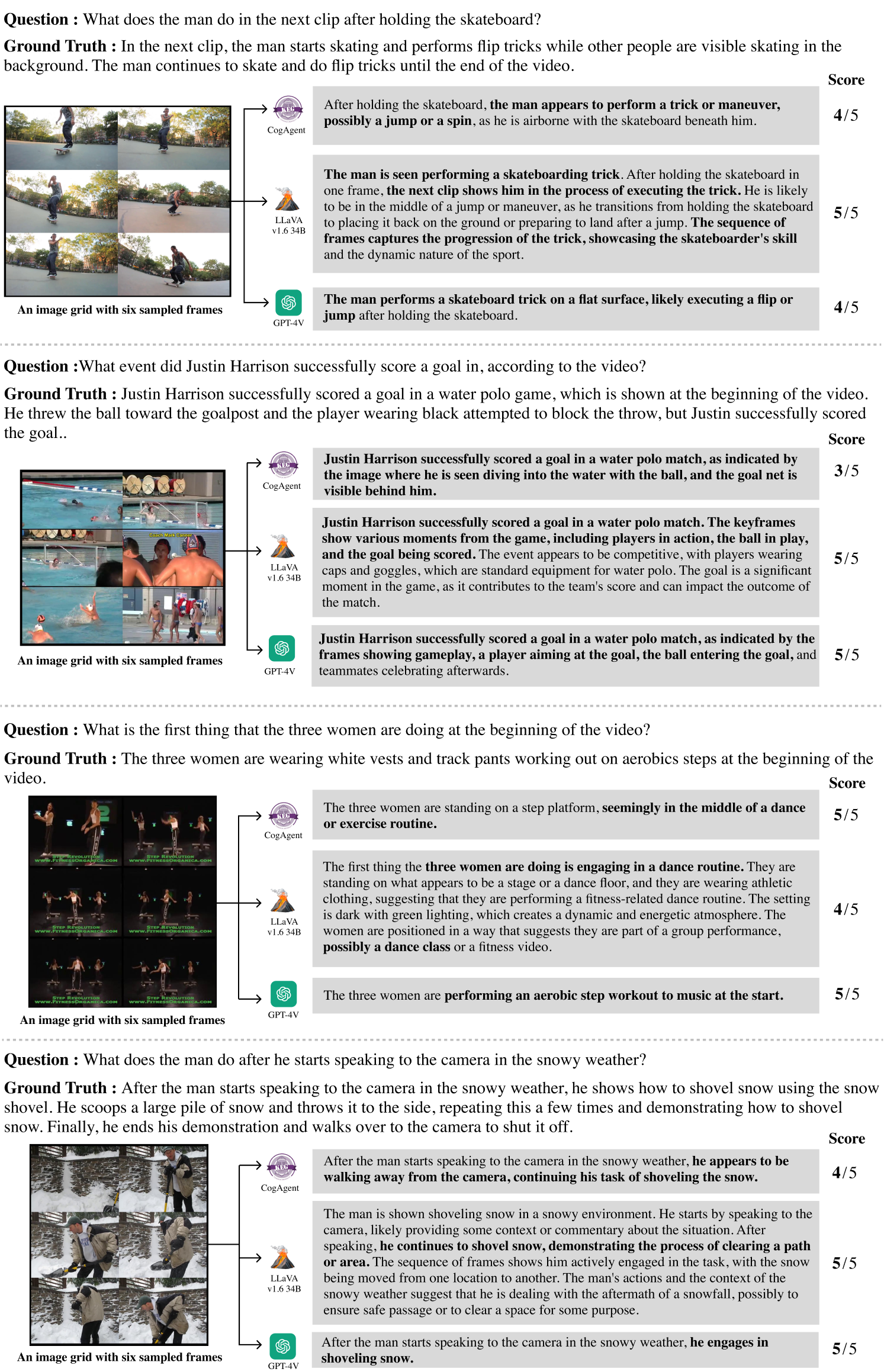}
  }
    \caption{Successful case examples of zero-shot VQA on Text Generation Performance benchmark.}
    \label{fig:success_example_open2}
\end{figure}

\begin{figure}[]
  \centering
   \scalebox{1.0}{
  \includegraphics[width=\linewidth]{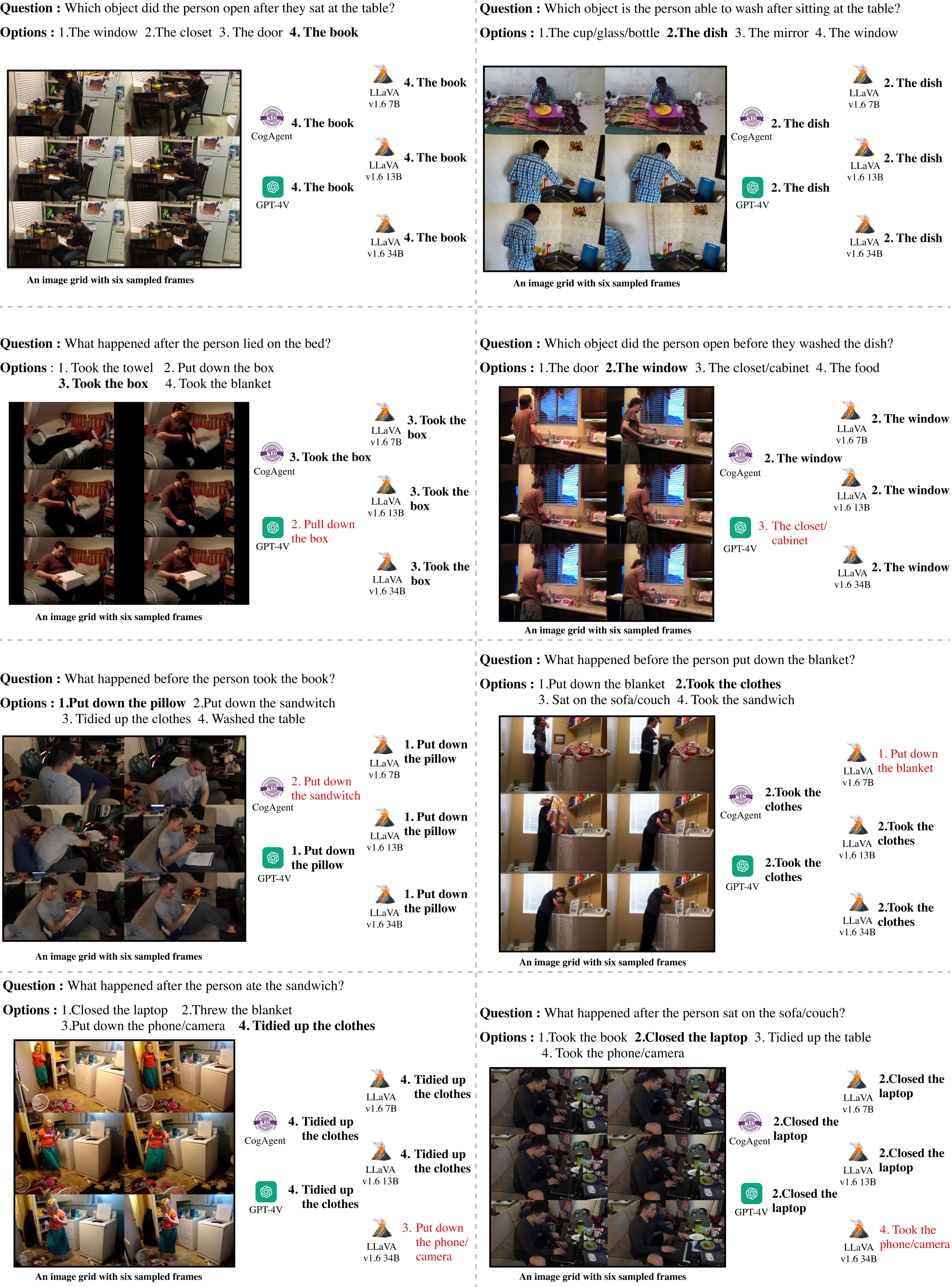}
  }
    \caption{Successful case examples of zero-shot VQA on STAR. Text highlighted in bold indicates the correct answer, while red text denotes an incorrect answer.}
    \label{fig:success_example_multiple}
\end{figure}

In this section, we present instances where IG-VLM provided successful responses for both open-ended and multiple-choice VQA. For open-ended VQA, \cref{fig:success_example_open1} and \cref{fig:success_example_open2} present the outcomes from three VLM models. For multiple-choice VQA, we present the results from five VLMs, as detailed in \Cref{fig:success_example_multiple}. In both open-ended and multiple-choice VQA, we present examples that not only highlight instances where all VLMs provided perfect answers but also include cases where some VLMs underperformed relative to the others.

\newpage
\FloatBarrier
\section{Failure Case Examples of VQA by IG-VLM}

\begin{figure}[]
  \centering
   \scalebox{0.96}{
  \includegraphics[width=\linewidth]{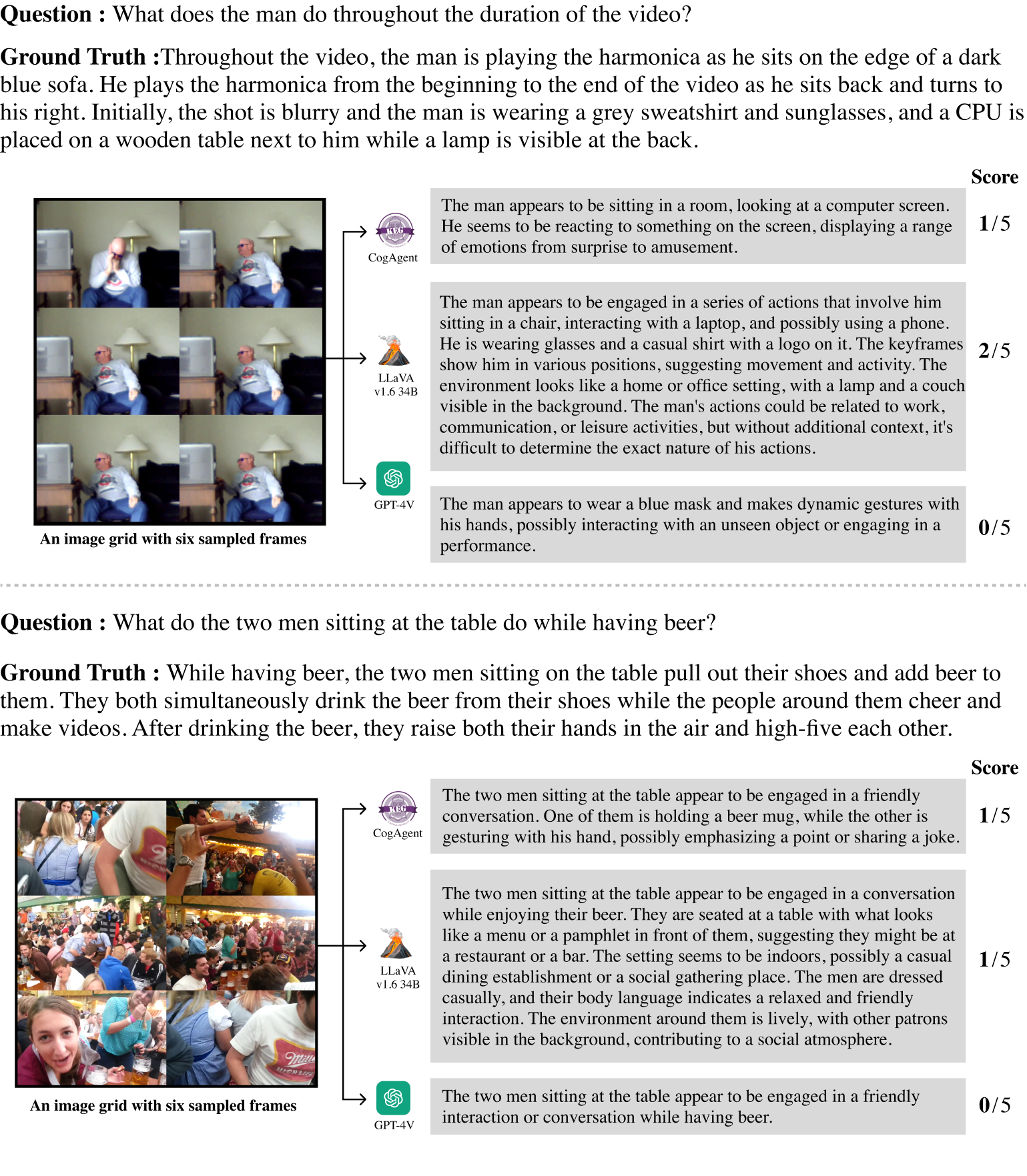}
  }
\caption{Failure case examples of zero-shot VQA on Text Generation Performance benchmark.}
\label{fig:fail_example_open}
\end{figure}

\begin{figure}[]
  \centering
   \scalebox{0.99}{
  \includegraphics[width=\linewidth]{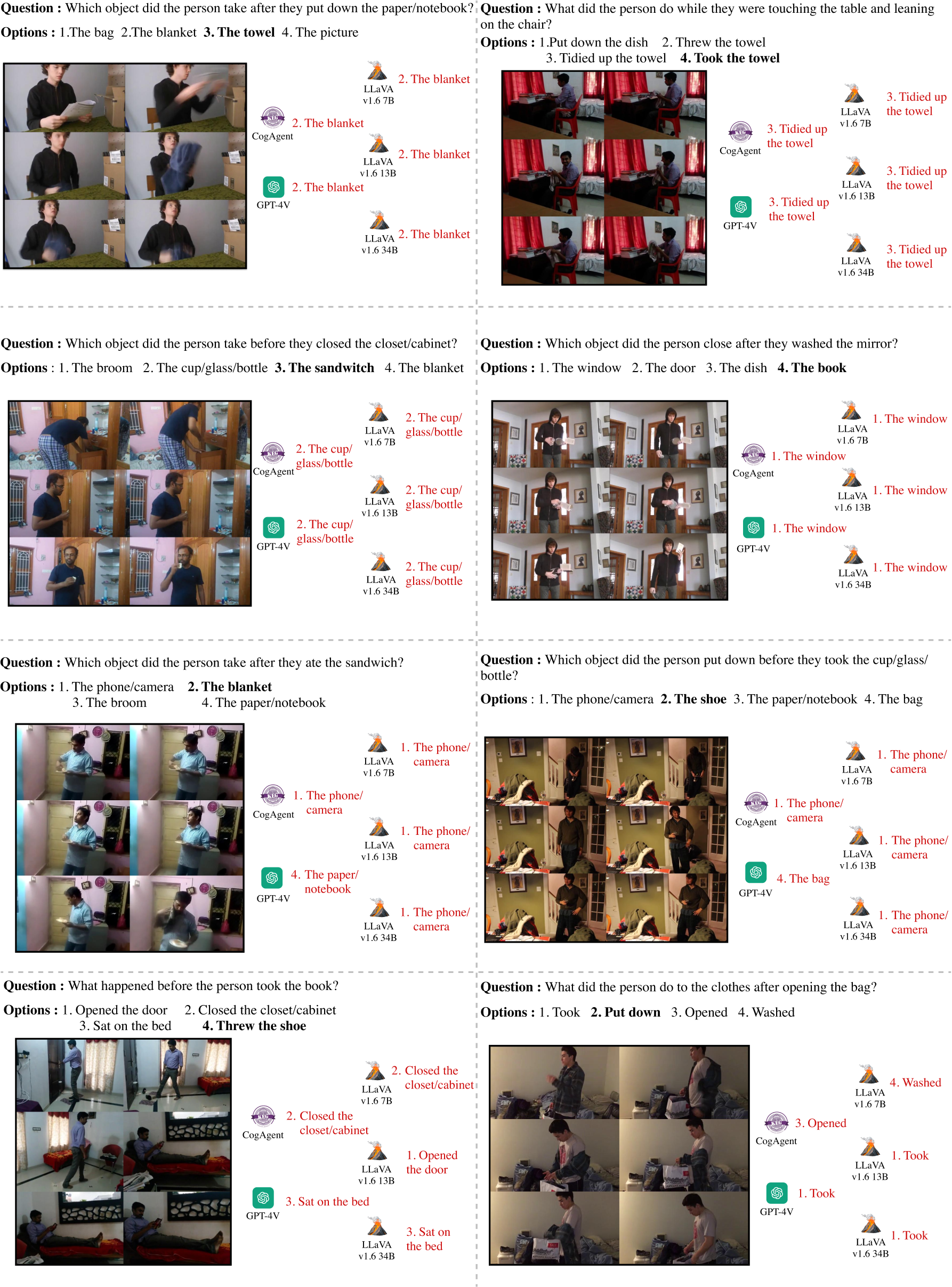}
  }
\caption{Failure case examples of zero-shot VQA on STAR. Text highlighted in bold indicates the correct answer, while red text denotes an incorrect answer.}
\label{fig:fail_example_multiple}
\end{figure}

In this section, we present instances where IG-VLM provided inaccurate responses in various VQA benchmarks. For open-ended VQA, we illustrate outcomes from three distinct VLMs in \cref{fig:fail_example_open}. In the context of multiple-choice VQA, results from five different VLMs are depicted in \cref{fig:fail_example_multiple}.
We identified three main reasons for failures: ambiguous question or ground truths, IG-VLM's limitation of six frames impacting its efficacy, and incorrect image object recognition by the VLMs using IG-VLM.

\newpage

\newpage
\FloatBarrier
\section{Experiment Corresponding to the Discussion on Multi-step Prompts}
The experimental results on \cref{tab:prompt_multiturn_ablation} show the performance for adopting multi-step reasoning prompts with IG-VLM. To ensure a fair comparison between single-step and multi-step reasoning, only grid guidance prompt was consistently used in all experiments.

\label{appendix:CoT_prompt_result}
\begin{table}[htbp!]
\centering
\caption{Experiment of multi-step prompt methods on manual prompt efficiency across ActivityNet-QA and NExT-QA.}
\label{tab:prompt_multiturn_ablation}
\scalebox{0.9}{
\begin{tabular}{c|l|cc|cccc}
\toprule

\multirow{2}{*}{\textbf{LLM Size}} & \multicolumn{1}{c|}{\multirow{2}{*}{\textbf{Method}}} & \multicolumn{2}{c|}{\textbf{ActivityNet-QA}} & \multicolumn{4}{c}{\textbf{NExT-QA}} \\
 & \multicolumn{1}{c|}{} & \textbf{ Accuracy } & \textbf{ Score } & \textbf{Cas.} & \textbf{Tem.} & \textbf{Des.} & \textbf{ Avg. Acc. } \\
\midrule
\multirow{5}{*}{7B} & Grid Prompt~(single-step) & 53.0 & 3.42 & 63.1 & 57.3 & 74.9 & \textbf{63.1} \\
& Zero-shot CoT\cite{kojima2022large}   & 51.6 & 3.42 & 58.6 & 50.1 & 69.6 & 57.6 \\
& Self-Consistency\cite{wang2023selfconsistency}            & \textbf{58.9} & 3.64 & 58.7 & 52.0 & 66.2 & 57.7 \\
& Plan and Solve\cite{Wang2023PlanandSolvePI}               & 51.1 & 3.41 & 61.8 & 53.9 & 68.9 & 60.4 \\
& Describe and Answer                                       & 51.0 & 3.36 & 62.0 & 57.7 & 75.9 & 62.8 \\ \midrule
\multirow{5}{*}{13B} &  Grid Prompt~(single-step)  & \textbf{54.2} & 3.53 & 61.6 & 55.7 & 70.8 & 61.2 \\
& Zero-shot CoT~\cite{kojima2022large} & 26.9 & 3.42 & 59.8 & 50.0 & 66.3 & 57.7 \\
& Self-Consistency\cite{wang2023selfconsistency}            & 27.0 & 3.75 & 60.7 & 51.6 & 68.6 & 59.0 \\
& Plan and Solve\cite{Wang2023PlanandSolvePI}               & 52.5 & 3.45 & 65.4 & 55.8 & 72.5 & \textbf{63.4} \\
& Describe and Answer                                       & 19.0 & 3.60 & 58.7 & 53.3 & 72.3 & 59.1 \\ \midrule
\multirow{5}{*}{34B}  & Grid Prompt~(single-step)   & \textbf{57.8} & 3.66 & 72.2 & 65.7 & 77.3 & 70.9 \\
& Self-Consistency\cite{wang2023selfconsistency}            & 54.5 & 3.40 & 23.4 & 23.4 & 23.0 & 23.3 \\
& Plan and Solve\cite{Wang2023PlanandSolvePI}               & 54.8 & 3.49 & 69.7 & 58.1 & 74.1 & 66.6 \\
& Describe and Answer                                       & 57.2 & 3.64 & 71.9 & 66.9 & 78.3 & \textbf{71.3} \\
\bottomrule
\end{tabular}
}
\end{table}

\FloatBarrier

\newpage
\section{Multi-step Prompts Templates}
\label{appendix:CoT_prompt_template}

\begin{figure}[]
  \centering
  \scalebox{0.80}{
  \includegraphics[width=\linewidth]{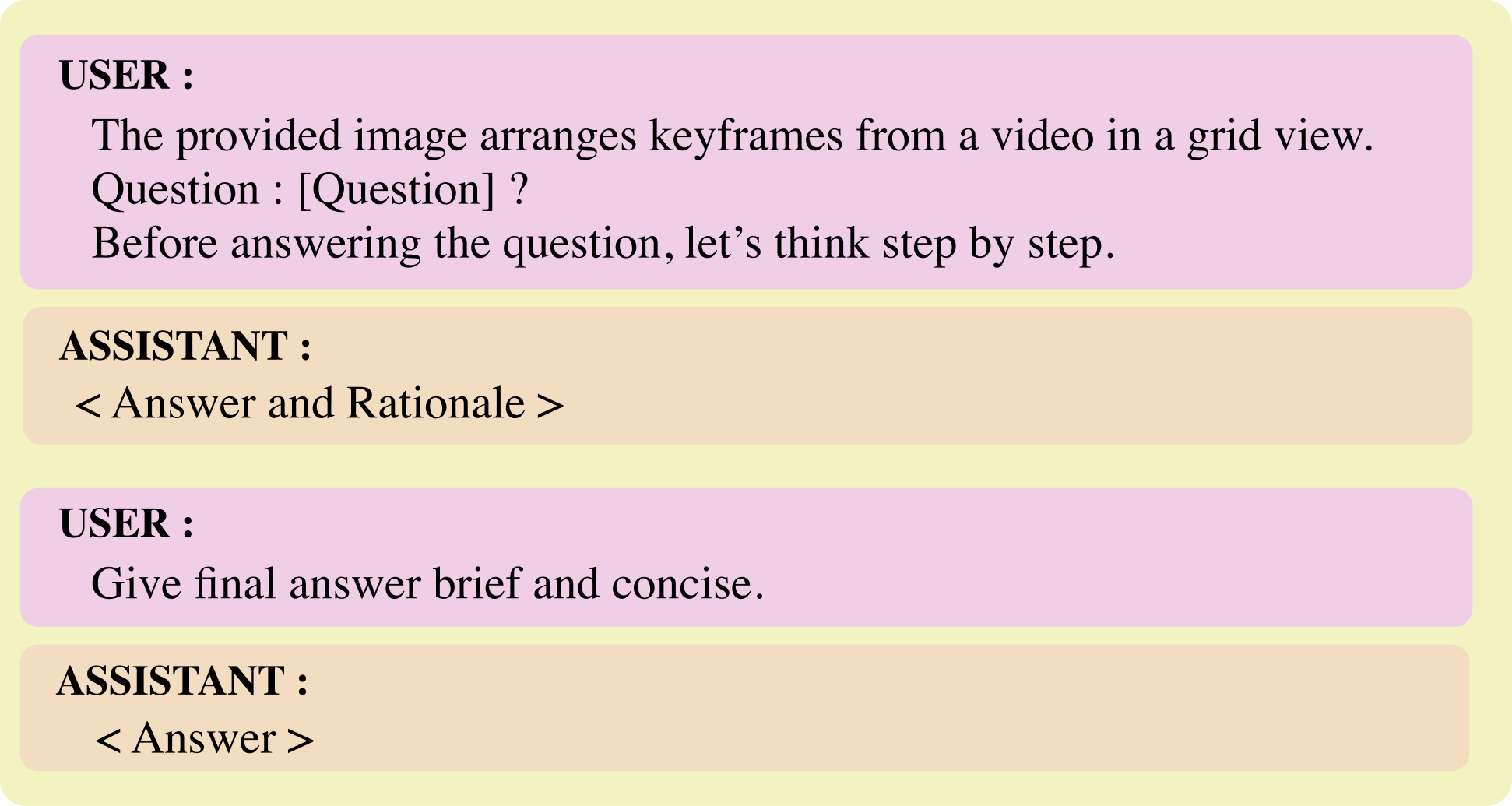}
  }
  \caption{Prompt for Zero-shot CoT}
  \label{fig:prompt_zeroshot_cot}
  \vspace{-30pt}
\end{figure}

\begin{figure}[]
  \centering
  \scalebox{0.80}{
  \includegraphics[width=\linewidth]{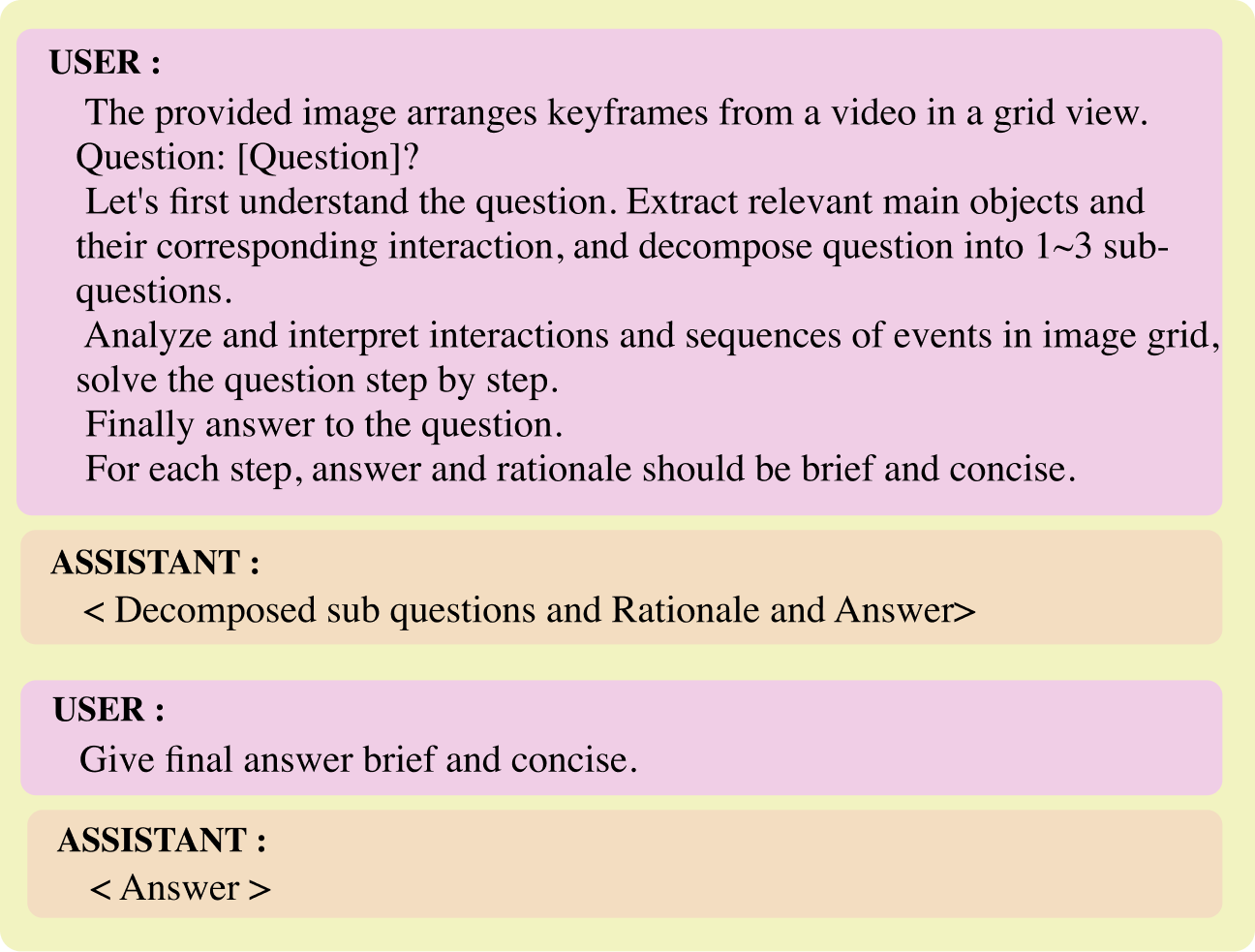}
  }
  \caption{Prompt for Plan and Solve}
  \label{fig:prompt_selfconsistency}
\end{figure}

\begin{figure}[]
  \centering
  \scalebox{0.80}{
  \includegraphics[width=\linewidth]{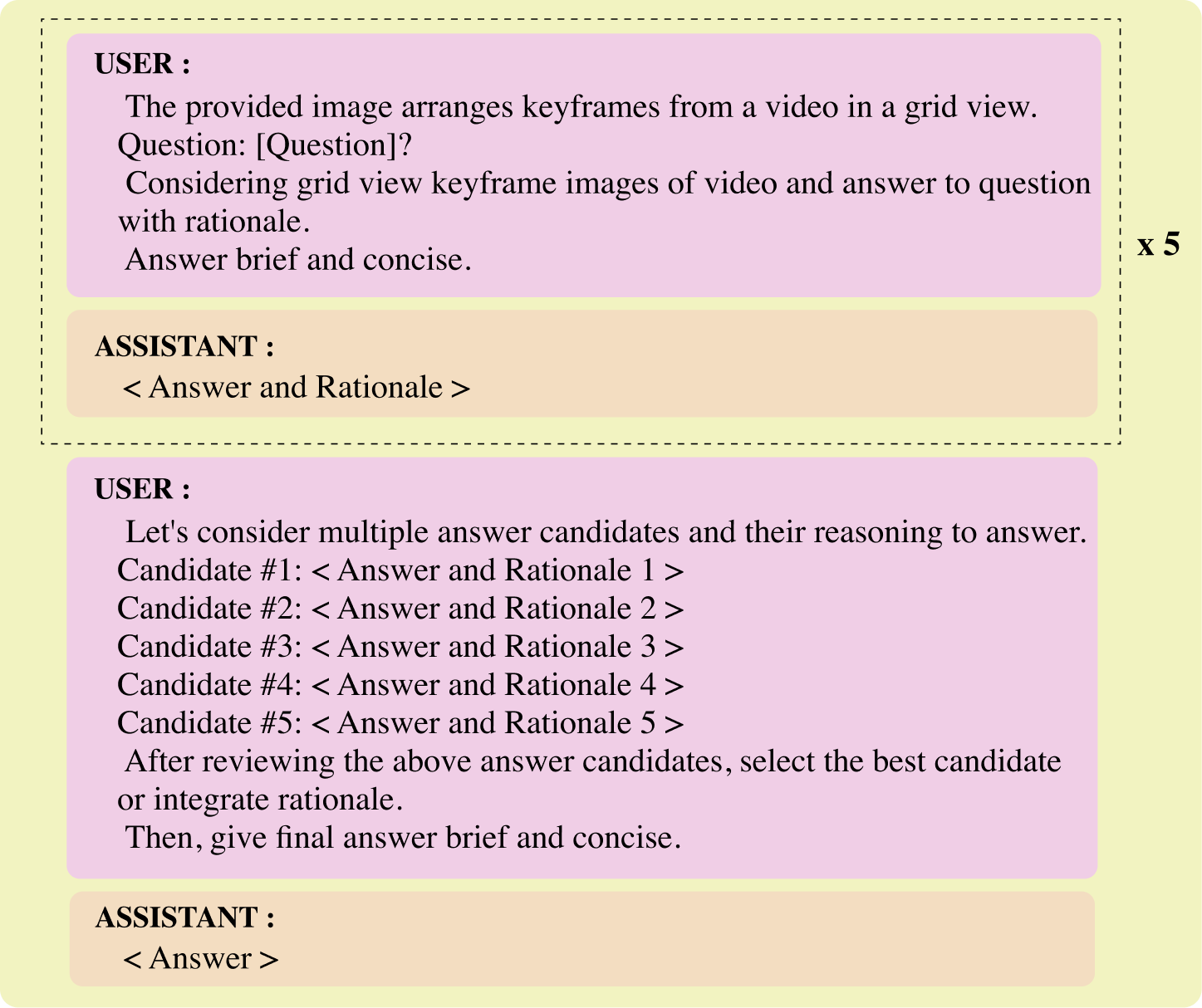}
  }
  \caption{Prompt for Self-Consistency}
  \label{fig:prompt_plan_and_solve}
\end{figure}

\begin{figure}[]
  \centering
  \scalebox{0.80}{
  \includegraphics[width=\linewidth]{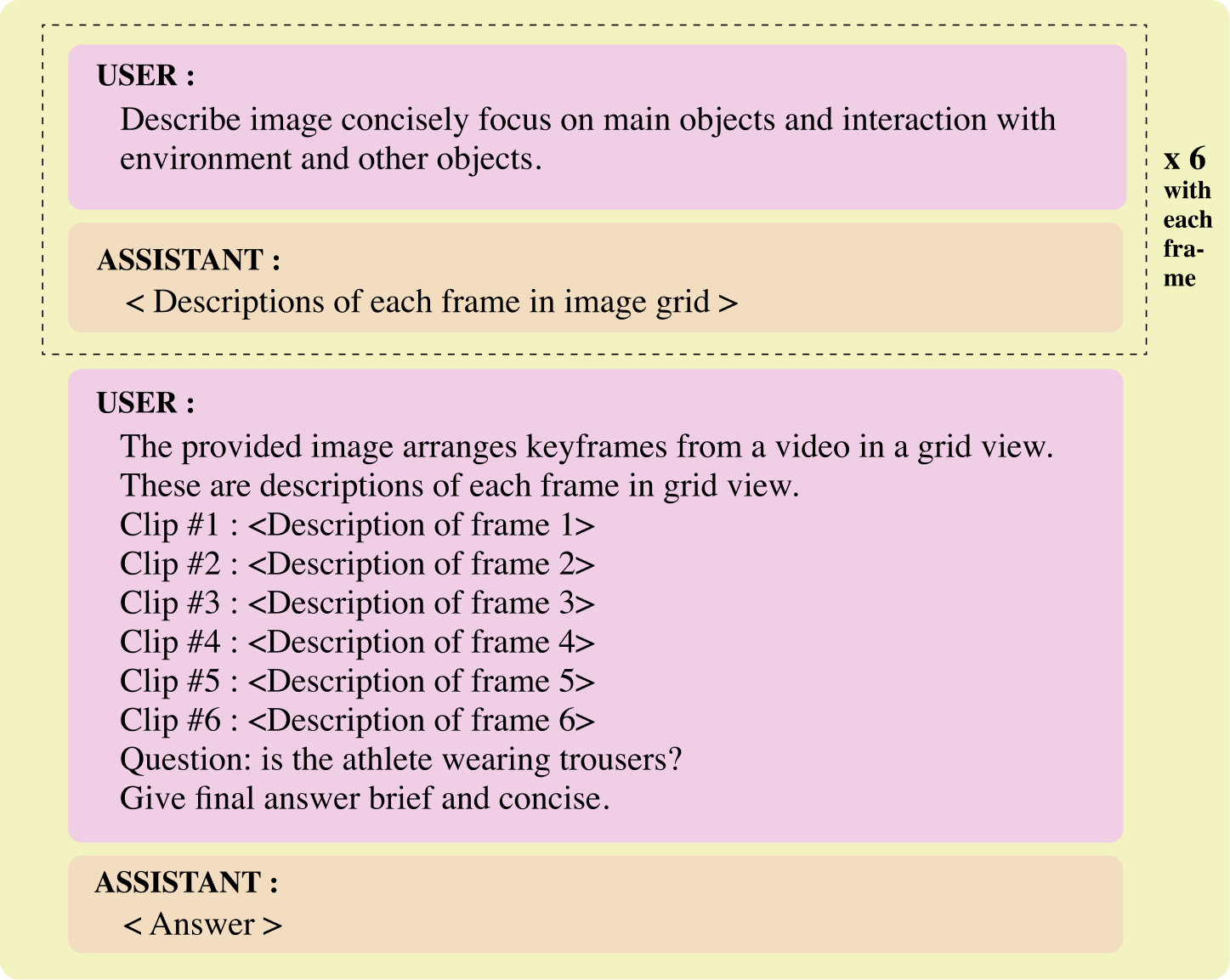}
  }
  \caption{Prompt for Describe and Answer}
  \label{fig:prompt_describe_and_answer}
\end{figure}

We present the actual templates of the multi-step prompts used in experiments with IG-VLM on \cref{fig:prompt_zeroshot_cot}, \cref{fig:prompt_selfconsistency}, \cref{fig:prompt_plan_and_solve}, and \cref{fig:prompt_describe_and_answer}.

\end{document}